\documentclass[letterpaper]{article} % DO NOT CHANGE THIS
\usepackage{aaai23} 
\usepackage{times}  % DO NOT CHANGE THIS
\usepackage{helvet}  % DO NOT CHANGE THIS
\usepackage{courier}  % DO NOT CHANGE THIS
\usepackage[hyphens]{url}  % DO NOT CHANGE THIS
\usepackage{graphicx} % DO NOT CHANGE THIS
\usepackage{natbib}  % DO NOT CHANGE THIS AND DO NOT ADD ANY OPTIONS TO IT
\usepackage{caption} % DO NOT CHANGE THIS AND DO NOT ADD ANY OPTIONS TO IT
\usepackage{newfloat}
\usepackage{listings}
\DeclareCaptionStyle{ruled}{labelfont=normalfont,labelsep=colon,strut=off} % DO NOT CHANGE THIS
\usepackage{xcolor}         % colors
\usepackage[ruled,vlined, linesnumbered]{algorithm2e}
\usepackage{adjustbox}
\usepackage{url}
\usepackage{multirow}
\usepackage{fancyhdr}
\usepackage{booktabs,subcaption,amsfonts,dcolumn}
\usepackage{physics,amsmath}
\usepackage{tabularx}

\newcommand{\our}{\emph{Frido}\xspace}  % TODO model name
\newcommand{\ourunet}{PyU-Net\xspace}  % TODO model name

\newcommand{\ul}{\underline}
\usepackage{tikz}

\definecolor{mygreen}{RGB}{130,237,155}

\makeatletter
\def\mfontsize{\f@size}

\makeatother

\nocopyright % -- Your paper will not be published if you use this command
\title{\our: Feature Pyramid Diffusion for Complex Scene Image Synthesis}

\author {
    Wan-Cyuan Fan\textsuperscript{\rm 1}\footnote{work done during research internship at Microsoft},
    Yen-Chun Chen\textsuperscript{\rm 2}\footnote{project lead}, \\
    Dongdong Chen\textsuperscript{\rm 2},
    Yu Cheng\textsuperscript{\rm 2},
    Lu Yuan\textsuperscript{\rm 2},
    Yu-Chiang Frank Wang\textsuperscript{\rm 1, 3}
}
\affiliations {
    \textsuperscript{\rm 1}National Taiwan University \ 
    \textsuperscript{\rm 2}Microsoft Corporation \ 
    \textsuperscript{\rm 3}NVIDIA\\
    \texttt{\small r09942092@ntu.edu.tw}\ \ \ \texttt{\small frankwang@nvidia.com}\\
    \texttt{\small\{yen-chun.chen, dochen, yu.cheng, luyuan\}@microsoft.com}
}

\usepackage{bibentry}

\begin{document}

\maketitle

\begin{abstract}
  
Diffusion models (DMs) have shown great potential for high-quality image synthesis. However, when it comes to producing images with complex scenes, how to properly describe both image global structures and object details remains a challenging task. In this paper, we present \our, a \textbf{F}eatu\textbf{r}e Pyram\textbf{i}d \textbf{D}iffusi\textbf{o}n model performing a multi-scale coarse-to-fine denoising process for image synthesis. Our model decomposes an input image into scale-dependent vector quantized features, followed by a coarse-to-fine modulation for producing image output. During the above multi-scale representation learning stage, additional input conditions like text, scene graph, or image layout can be further exploited. Thus, \our can be also applied for conditional or cross-modality image synthesis. We conduct extensive experiments over various unconditioned and conditional image generation tasks, ranging from text-to-image synthesis, layout-to-image, scene-graph-to-image, to label-to-image. More specifically, we achieved state-of-the-art FID scores on five benchmarks, namely layout-to-image on COCO and OpenImages, scene-graph-to-image on COCO and Visual Genome, and label-to-image on COCO.\footnote{Code is available at \url{https://github.com/davidhalladay/Frido}.}

\end{abstract}

\section{Introduction}
\label{sec:intro}

\begin{figure}[t]
  \centering
  \includegraphics[page=11,trim={167 100 167 80}, clip, width=0.45\textwidth]{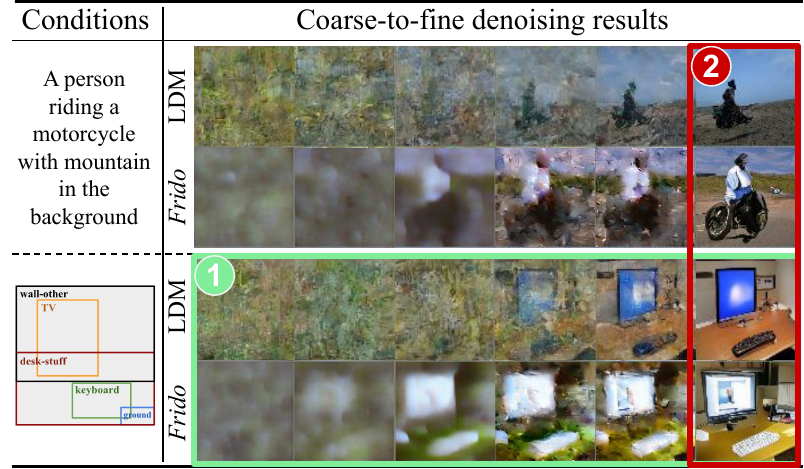}
  \vspace{-2mm}
  \caption{Illustration of \our. Given a cross-modal condition, \our generates images in \textcolor{mygreen}{\textcircled{1}} a coarse-to-fine manner from structure to object details, producing outputs with \textcolor{red}{\textcircled{2}} high semantic correctness and quality. Note that existing models such as the LDMs are not designed to distinguish between high/low-level visual information.}
  \vspace{-4mm}
  \label{demo:teaser}
\end{figure}

Generating photo-realistic images is a critical task in computer vision research. In this task, a generative model is designed to learn the underlying data distribution of a given set of images and to be capable of synthesizing new samples from the learned distribution. To this end, series of methods were proposed, including VAEs~\cite{kingma2013auto, van2017neural}, GANs~\cite{goodfellow2014generative, radford2015unsupervised}, flow-based methods~\cite{dinh2014nice, kingma2018glow}, and the trending diffusion models~(DMs)~\cite{sohl2015deep, ho2020denoising}. The quality of the generated images has been improved rapidly with the contribution of these lines of works. Moreover, the task itself also evolves from object-centric image synthesis without conditions to complex scene image generation, and sometimes based on multi-modal conditions (e.g., texts, layouts, labels, and scene-graphs).

Recently, diffusion models~\cite{ho2020denoising, nichol2021improved, ho2022cascaded, rombach2022high, ramesh2022hierarchical} have demonstrated a remarkable capability of high-quality image synthesis and outperform other classes of generative approaches on multiple tasks, including but not limited to unconditional image generation, text-to-image generation, and image super-resolution.
Despite the encouraging progress, diffusion models may fall short when targeted images are more complex and conditioning inputs are highly abstractive. The composition of objects and parts, along with high-level semantic relations are prevailing in those tasks, which are less seen in earlier object-centric benchmarks and may be essential to higher quality generation.

\begin{figure*}[t]
  \centering
  \includegraphics[page=4,trim={690 1002 695 1005}, clip, width=0.97\textwidth]{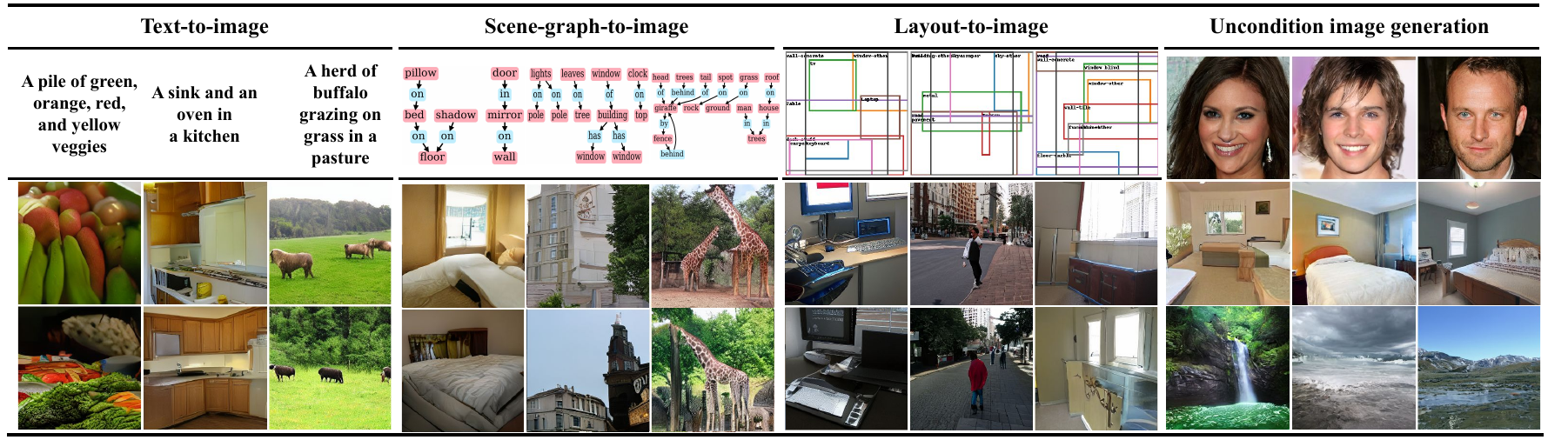}
  \vspace{-3mm}
  \caption{Generated examples of \our on various tasks. From left to right, we show the examples of text-to-image (T2I) on COCO 2014, scene-graph-to-image (SG2I) on Visual Genome, layout-to-image (Layout2I) on COCO-stuff, unconditional image generation on Landscape, CelebA, LSUN-bed. All images are 256x256 resolution. Note that, for conditional generation, we adopt classifier-free guidance with s=1.5 while testing. Please refer to supplementary for full-scale version.}
  \vspace{-5mm}
  \label{demo:big}
\end{figure*}

In particular, we point out two major challenges in existing DM works. \textit{First}, most of existing DMs deal with feature maps or image pixels at a single scale/resolution, which might not be able to capture image semantics or compositions in real-world complex scenes. Take the first row of Figure~\ref{demo:teaser} as examples, it can be seen that while LDM~\cite{rombach2022high} generates images containing a ``person" given the text condition, semantic structures of ``riding a motorcycle" and ``mountain in the background" are not sufficiently produced. \textit{Second}, expensive computational resources are typically required for DMs during training and testing due to the iterative denoising processes, especially for producing high-resolution outputs. This not only limits the accessibility but also results in massive carbon emissions. Therefore, a computationally efficient diffusion model that leverages coarse/high-level synthesized outputs for introducing multi-scale visual information would be desirable.

To address these limitations, we propose \textbf{\our}, a \textbf{F}eatu\textbf{r}e Pyram\textbf{i}d \textbf{D}iffusi\textbf{o}n model for complex scene image generation.\footnote{\our is pronounced as ``free-dow".} \our is a novel multi-scale coarse-to-fine diffusion and denoising framework, which allows synthesizing images with enhanced global structural fidelity and realistic object details.
Specifically, we introduce a novel \emph{feature pyramid U-Net}~(\ourunet) with a \emph{coarse-to-fine modulation} design, enabling our model to denoise visual features from multiple spatial scales in a top-down fashion.
These multi-scale features are produced by our MS-VQGAN, a newly designed multi-scale variant of VQGAN~\cite{esser2021taming} that encodes images into multi-scale visual features (discrete latent codes).
As can be seen in Figure~\ref{demo:teaser}, as the feature gradually being denoised, the images are reconstructed in a coarse-to-fine manner (decoded by our MS-VQGAN decoder), from global structures to fine-grained details.
On the other hand, a recent competitive diffusion model~\cite{rombach2022high} reconstructs images uniformly across spatial scales.

\our is a generic diffusion framework that can synthesize images from diverse, multi-modal inputs, including texts, box-layouts, scene-graphs, and labels.
Moreover, our model introduces minimal extra parameters while allowing us to speed up the notoriously slow inference of conventional DMs.
Extensive experiments are done to demonstrate the effectiveness of the new designs.
Our contributions are summarized as follows.
($i$) We propose \our, a novel diffusion model to generate photo-realistic images from multi-modal inputs, with a \emph{coarse-to-fine} prior that is under-explored in the DM paradigm.
($ii$) Empirically, we achieve \emph{5 new state-of-the-art} results, including layout-to-image on COCO and OpenImages, scene-graph-to-image on COCO and Visual Genome, and label-to-image on COCO, all are complex scenes with highly abstractive conditions.
($iii$) In practice, \our \emph{inferences fast}, shown by a head-to-head comparison with an already fast diffusion model, the LDM.

\section{Preliminary}
\label{sec:related}

Multiple lines of works to generate photo-realistic images have been proposed, including VAEs, GANs, and Invertible-Flows, and achieved impressive results for object-centric images. However, VAEs suffer from blurry outputs. GANs are notoriously hard to train and lack diversity. Flow-based model suffers shape distortions due to imperfect inverse transform.
Our work belongs to the paradigm of diffusion models (DMs), which have been shown to best synthesize high quality images among all deep generative methods.
For completeness, we summarize the fundamentals of DMs and a recent improvement, Latent Diffusion Models~(LDMs)~\cite{rombach2022high}.

\paragraph{Diffusion Models for Image Generation}
A diffusion model (DM) contains two stages: forward (diffusion) and backward (denoising) processes. In the forward process, the given data $\vectorbold{x}_0 \sim q(\vectorbold{x}_0)$ is gradually destroyed into an approximately standard normal distribution $\vectorbold{x}_T \sim p(\vectorbold{x}_T)$ over $T$ steps, where $q$ and $p$ denote the given data manifold and the standard Gaussian distribution, respectively; and $\vectorbold{x}$ denotes a data point from $q$. The diffusion process, formulated by \citet{ho2020denoising}, are shown as follows:
\begin{equation}
    \begin{aligned}
        q(\vectorbold{x}_{1:T}|\vectorbold{x}_0)
        &=\prod\limits_{t=1}^{T}q(\vectorbold{x}_t|\vectorbold{x}_{t-1}),\text{ and } \\
        q(\vectorbold{x}_t|\vectorbold{x}_{t-1})&=\mathcal{N}(\sqrt{1-\beta_{t}}\vectorbold{x}_{t-1}, \beta_t \vectorbold{I}).
    \end{aligned}
    \label{eq:diffusion}
\end{equation}
, where $\beta$ denotes the noise schedule. Can be fixed or learned. By reversing the forward process, \citet{ho2020denoising} obtained the backward process:
\begin{equation}
    \begin{aligned}
        p_\theta(\vectorbold{x}_{0:T})&=p(\vectorbold{x}_{T})\prod\limits_{t=1}^{T}p_\theta(\vectorbold{x}_{t-1}|\vectorbold{x}_{t}),\text{ and } \\  p_\theta(\vectorbold{x}_{t-1}|\vectorbold{x}_{t})&=\mathcal{N}(\vectorbold{x}_{t-1};\mu_\theta(\vectorbold{x}_t, t), \sigma_\theta(\vectorbold{x_t}, t)).
    \end{aligned}
    \label{eq:denoising}
\end{equation}
, where $\theta$ denotes the learnable parameters, a U-Net~\cite{RonnebergerFB15} in \citet{ho2020denoising}. This is implemented by a neural network predicting each of the denoising steps;
and it can be viewed as a Markov chain with a learned Gaussian transition distribution~\cite{dhariwal2021diffusion, pandey2022diffusevae}.

In practice, we randomly sample a timestep $t$ in $[0, T]$, and then compute $\vectorbold{x}_t$ by interpolating $\vectorbold{x}_0$ and $\epsilon$ with the weight schedule $\beta_t$, where $\epsilon$ is sampled Gaussian noise.
The denoising network $\epsilon_{\theta}$ is trained by the following loss:
\begin{equation}
    \begin{aligned}
        \mathcal{L}_{DM} = \mathbb{E}_{\vectorbold{x}_0, \epsilon, t} \left[ \|\epsilon-\epsilon_\theta(\vectorbold{x}_t)\|^2 \right].
    \end{aligned}
    \label{eq:ddpm_objtive}
\end{equation}
At a higher level, this loss trains the network to predict the step noise $\epsilon$ applied on $\vectorbold{x}_{t-1}$ given $\vectorbold{x}_t$.
To synthesize an image, one can run this denoising network for $T$ steps to gradually denoise a random noise image.

\paragraph{Latent Diffusion Models} Most DMs~\cite{nichol2021glide, dhariwal2021diffusion} operate on the original image pixels, yielding high dimensional data manifold with input $\vectorbold{x}_0 \in \mathcal{R}^{3 \times H \times W}$.
Such high-dimensional inputs cost huge computation for the diffusion and denoising processes at both training and inference. Very recently, Latent Diffusion Models~(LDMs)~\cite{rombach2022high} are proposed to adopt DMs to learn the low-dimensional latent codes, encoded by a VQGAN~\cite{esser2021taming} or KL-autoencoder~\cite{rombach2022high}. Given an image $\vectorbold{x}_0$ and the pre-trained autoencoder, containing encoder $\mathcal{E}$ and decoder $\mathcal{D}$, the corresponding latent codes $\vectorbold{z_0}=\mathcal{E}(\vectorbold{x_0})$ can be produced, where $\vectorbold{z}_0 \in \mathcal{R}^{c \times h \times w}$, $c$ is usually set to $4$; and $h, w$ are downsampled $8-16$ times from $H, W$.
By replacing the image data point $\vectorbold{x}$ in Eq.~\eqref{eq:diffusion} and Eq.~\eqref{eq:denoising} with the encoded latent $\vectorbold{z}$, the diffusion and denoising processes of a LDM can be derived:
\begin{equation}
    \begin{aligned}
        q(\vectorbold{z}_{1:T}|\vectorbold{z}_0)&=\prod\limits_{t=1}^{T}q(\vectorbold{z}_t|\vectorbold{z}_{t-1}), \text{ and }\\
        p_\theta(\vectorbold{z}_{0:T})&=p(\vectorbold{z}_{T})\prod\limits_{t=1}^{T}p_\theta(\vectorbold{z}_{t-1}|\vectorbold{z}_{t}).
    \end{aligned}
    \label{eq:ldm}
\end{equation}
At inference, the final output image can be reconstructed from the denoised latent $\tilde{\vectorbold{x}}_0 = \mathcal{D}(\tilde{\vectorbold{z}}_0)$, where $\tilde{\vectorbold{z}}_0$ is sampled and denoised using Eq.~\eqref{eq:ldm}.
Since $T$ is typically set to $500-1000$ in practice, and the autoencoding is a one-time operation per image, the overall computation is greatly reduced due to the much lower resolution of $\vectorbold{z}_0$.

\section{Methodology}
\label{sec:method}

\begin{figure*}
     \centering
     \vspace{-1mm}
     \begin{subfigure}[b]{0.33\textwidth}
         \centering
         \includegraphics[page=2,trim={225 2050 1230 108}, clip, width=0.95\textwidth]{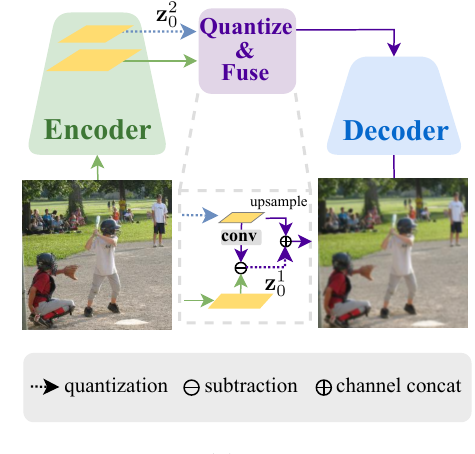}
         \vspace{-2mm}
         \caption{Architecture of MS-VQGAN.}
         \label{subfig:ms-vqgan}
     \end{subfigure}
     \hfill
     \begin{subfigure}[b]{0.65\textwidth}
         \centering
         \includegraphics[page=2,trim={460 2050 750 97}, clip, width=0.95\textwidth]{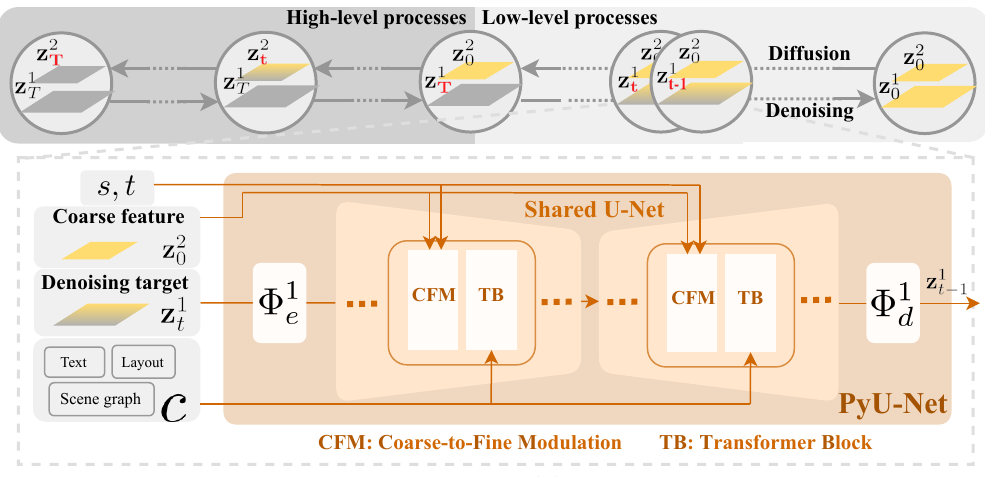}
         \vspace{-2mm}
         \caption{Details of the diffusion and denoising processes.}
         \label{subfig:our-diffusion}
     \end{subfigure}
     \hfill
     \vspace{-2mm}
    \caption{Overview of \our (best viewed in color).
         How MS-VQGAN encodes an image into multi-scale feature maps $\vectorbold{z}^1_0, \vectorbold{z}^2_0$ is illustrated in (a). The quantization enables VQ-VAE learning; and the fusion allows merging all representations from high to low level for the decoder to reconstruct an image.
         The upper half of (b) demonstrate the coarse-to-fine process, where the denoising is completed for high-level first, and then the lower one. The lower half of (b) details each denoising step. A U-Net is shared not only across timestep $t$ but also the scale level $s$. Coarse-to-fine gating will be explained in Figure~\ref{arch_2}.
         }
    \label{arch_1}
    \vspace{-4mm}
\end{figure*}

Although existing DMs generate high-resolution images for a single object with outstanding quality, most of them only deal with feature maps or image pixels at a single resolution. Since they treat high and low-level visual concepts equally, it is not easy for such DM models to describe the corresponding image semantics or composition. This might limit their uses for synthesizing complex scene images.

To enhance DMs with global structural modeling, we propose to model the latent features in a coarse-to-fine fashion via feature pyramids. We first introduce the Multi-Scale Vector Quantization model~(MS-VQGAN), which encodes the image into latent codes at several spatial levels.
Next, we propose the feature pyramid diffusion model~(\our), extending the diffusion and denoising into a multi-scale, coarse-to-fine fashion. To achieve these, we design a new feature Pyramid U-Net~(\ourunet), equipped with a special modulation mechanism to allow coarse-to-fine learning. In this section, we introduce each component in detail.

\subsection{Learning Multi-Scale Perceptual Latents}
Before we model an image in a coarse-to-fine fashion, we first encode it into latent codes with several spatial resolutions. Extending from VQGAN~\cite{esser2021taming}, we train a multi-scale auto-encoder, named MS-VQGAN, with a feature pyramid encoder $\mathcal{E}$ and decoder $\mathcal{D}$.
As shown in Figure~\ref{subfig:ms-vqgan}, given an image $\vectorbold{x}_0$, the encoder $\mathcal{E}$ firstly produces a latent feature map set of $N$ scales $\mathcal{Z}= \mathcal{E}(\vectorbold{x}_0) = \{\vectorbold{z}^{1:N}\}$, where $\vectorbold{z}^t \in \mathcal{R}^{c \times \frac{s}{2^{t-1}} \times \frac{s}{2^{t-1}}}$. Note that $N$ and $c$ denote the number of feature maps (stages) and the channel size of the feature, respectively; and $s$ represents the size of the largest feature map.
In this design, we are encouraging $\vectorbold{z}^1$ to preserve lower-level visual details and $\vectorbold{z}^N$ to represent higher-level shape and structures.
Secondly, after quantizing and fusing, we upsample these features to the same shape, concat them, and feed them into the decoder $\mathcal{D}$ and reconstruct the image $\mathcal{D}(\mathcal{Z}) = \vectorbold{\tilde{x}_0}$.
The objective for this auto-encoder module is the weighted sum of $l_2$ loss between $\vectorbold{x}_0$ and $\vectorbold{\tilde{x}}_0$, and other perceptual losses\footnote{Patch discriminator loss and perceptual reconstruction loss.} in VQGAN.

We highlight that, with this design, MS-VQGAN can not only encode the input image into multi-scale codes of different semantic levels but also preserve more structure and detail, as later analyzed in Section~\ref{sec:model_ana} of model ablation.

\subsection{Feature Pyramid Latent Diffusion Model}
\label{method:frido}
After the MS-VQGAN is trained, we can use it to encode an image into multi-level feature maps $\mathcal{Z}$.
Next, we introduce the feature Pyramid Diffusion Model (\our) to model the underlying feature distribution and then generate images from noises. 
Similar to other DMs, \our contains two parts: the \textit{diffusion process} and the \textit{denoising process}.

\subsubsection{Diffusion Process of \our}
Instead of naively adding noises simultaneously on all $N$ feature scales $\mathcal{Z} = \{\vectorbold{z}^1, ..., \vectorbold{z}^N\}$ at each of the $T$ steps, we conduct diffusion process sequentially from low-level ($\vectorbold{z}^1$) to high-level ($\vectorbold{z}^N$), and each level takes $T$ diffusion steps (total of $N \times T$ timesteps). 
See the top half of Figure~\ref{subfig:our-diffusion} for an illustration.

Different from the classical diffusion process that corrupts pixels into noise in an unbiased way, we observe that \our's diffusion process starts from corrupting the object details, object shape, and finally the structure of the entire image. This allows \our to capture information in different semantic levels. See Fig~\ref{demo:teaser} for qualitative examples.

\subsubsection{Denoising Process of \our}
In the denoising phase, a sequence of neural function estimator $\epsilon_{\theta, t, n}$ is trained, where $t=1, 2, \dots, T$ and $n = N, N-1, \dots, 1$.
In order to denoise scale-by-scale, we introduce a novel feature pyramid U-Net~(\ourunet) as the neural approximator.
\ourunet can denoise the multi-scale features from high-level $\vectorbold{z}^N$ to low-level $\vectorbold{z}^1$ sequentially, achieving a coarse-to-fine generation. 
We highlight that, different from the LDMs, our \ourunet is more suitable for coarse-to-fine diffusion with these two novel features: (1) \emph{shared U-Net} with \emph{lightweight level-specific layers} that project features of different levels to a shared space so that the heavier U-Net can be reused across all levels, reducing the trainable parameters, and (2) \emph{coarse-to-fine modulation} to condition the denoising of low-level features on high-level ones that are already generated.

\paragraph{Feature Pyramid U-Net} 
The proposed \ourunet learns the denoising process in a coarse-to-fine fashion.
Take $N = 2$ $(\mathcal{Z} = \{\vectorbold{z}^1_0$, $\vectorbold{z}^2_0\})$ as an example (shown in Figure~\ref{arch_1}(b)),
\ourunet takes 4 inputs: (1) stage $s$ and timestep $t$ embeddings, (2) high-level feature conditions $\vectorbold{z}^2_0$, (3) target feature map $\vectorbold{z}^1_t$, and (4) other cross-modal conditions $\vectorbold{c}$.
By jointly observing these inputs, \ourunet predicts the noise $\epsilon$ applied on the target feature $\vectorbold{z}^1_t$, as shown in Figure~\ref{subfig:our-diffusion}.

Instead of using a separate U-Net for each stage $n$, we opt for a single shared U-Net to reduce the parameter count.
The input denoising target $\vectorbold{z}^1_t$ is first projected by level-specific layers $\Phi^{1}_e$ into a shared space so that a shared U-Net can be applied.
Finally, another level-specific projection $\Phi^{1}_d$ decodes the U-Net output to predict the noise $\epsilon$ added on $\vectorbold{z}^1_{t}$, with the following objective similar to Eq.~\eqref{eq:ddpm_objtive}:
\begin{equation}
    \begin{aligned}
        \mathcal{L}_{\text{\our}} = \mathbb{E}_{z^n_0, \epsilon, t}
        \left[ \|\epsilon-\epsilon_\theta(\vectorbold{z}^n_{t}, \vectorbold{z}^{n+1:N}_0, t)\|^2 \right] .
    \end{aligned}
\end{equation}
We note that \ourunet not only reduces the trainable parameters but also improves the results compared to vanilla per-stage U-Nets. For analysis, please refer to the experiments. Also, for training efficiency, we adopt the teacher forcing trick similar to sequence-to-sequence language models~\cite{brown2020language}, where ground truth feature conditions are used while denoising the low-level map.

\begin{figure}[t]
  \centering
  \includegraphics[page=2,trim={255 1750 1195 500}, clip, width=0.4\textwidth]{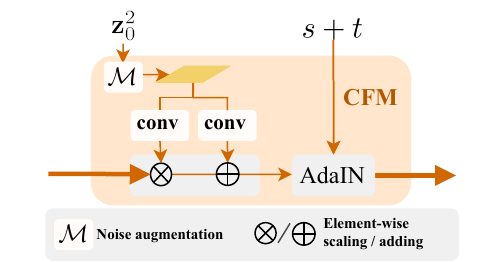}
  \vspace{-2mm}
  \caption{Framework of coarse-to-fine modulation in \ourunet. Note that we ignore some intermediate convolution layers and SiLU layers for simplification.}
  \vspace{-5mm}
  \label{arch_2}
\end{figure}

\paragraph{Coarse-to-Fine Modulation} \our produces the latent codes sequentially from high-level to low-level feature maps.
For example, while generating $\vectorbold{z}^1_t$ (low-level), the model is conditioned on $\vectorbold{z}^2_0$ (high-level).
We, therefore, introduce a coarse-to-fine modulation as shown in Figure~\ref{arch_2}.

Our coarse-to-fine modulation (CFM) is designed to incorporate (1) 2D high-level features, and (2) 1D stage and time embedding into residual blocks, allowing \our to have the high-level feature as well as stage-temporal awareness. Therefore, in our proposed CFM, there are two types of modulation conducted upon normalized features sequentially,  between which an extra convolution (conv) and SiLU layer~\cite{elfwing2018sigmoid} are inserted. 

Specifically, given the high-level ground truth $\vectorbold{z}^2_0$, we apply noise augmentation by
$\mathcal{M}(\vectorbold{z}^{2}_0)=f_z$, where $\mathcal{M}(\vectorbold{z}^{n}_0) = \alpha \cdot \vectorbold{z}^{n}_0 + (1-\alpha) \cdot \epsilon$. We note that $\epsilon \sim \mathcal{N}(\textbf{0}, \textbf{I})$, and the scaler $\alpha$ is a hyper-parameter.
After that, assume the input of CFM to be $f_i$, in the first modulation, we modulate the normalized feature norm($f_i$) with 2D scaling and shifting parameters from high-level feature $f_z$ with two convs respectively, producing intermediate representation $h$ as follows:
\begin{equation}
    \begin{aligned}
        h = \text{conv} \circ \text{SiLU} (\text{conv}(f_z)\times\text{norm}(f_i)+\text{conv}(f_z)).
    \end{aligned}
    \label{eq:cfm}
\end{equation}
In the second modulation, to equip U-Net with stage-temporal awareness, we further modulate $h$ with 1D stage+time embedding and produce output $f_o$, similar to Eq.~\ref{eq:cfm}. Note that we use linear layers to transform $s+t$ and also add a conv and SiLU after the AdaIN~\cite{huang2017arbitrary}.

To summarize, we highlight that our \ourunet framework equips DM with the ability to learn in a coarse-to-fine fashion with a moderate increase of parameters compared to classical hierarchical learning strategy~\cite{razavi2019generating}.
\our inherits three generative paradigms, VAE, GAN, and DM, and is further embedded with a coarse-to-fine prior.
Moreover, the diffusion operates on lower-resolution maps first, resulting a speedup at inference.
Later, we show that SOTA results can be achieved under a similar compute budget to a strong, fast DM.

\begin{table}[!tb]
\centering
\resizebox{0.48\textwidth}{!}{
\begin{tabular}{lccc}
\multicolumn{1}{l|}{Methods} & FID$\downarrow$ & IS$\uparrow$ & CLIP$\uparrow$ \\ \hline
\multicolumn{4}{c}{Methods under standard T2I setting} \\ \hline
\multicolumn{1}{l|}{AttnGAN~\cite{xu2018attngan}} & 33.10 & 23.61 & - \\
\multicolumn{1}{l|}{Obj-GAN~\cite{li2019object}} & 36.52 & 24.09 & - \\
\multicolumn{1}{l|}{DM-GAN~\cite{zhu2019dm}} & 27.34 & \textbf{32.32} & - \\

\multicolumn{1}{l|}{DF-GAN~\cite{tao2022df}} & 21.42 & - & - \\
\multicolumn{1}{l|}{LDM-8$^\dagger$~\cite{rombach2022high}} & 17.61 & 19.34 & 0.6500 \\
\multicolumn{1}{l|}{VQ-diffusion$^\ddag$~\cite{gu2022vector}} & 14.06 & 21.85 & 0.6770 \\
\multicolumn{1}{l|}{LDM-8-G$^\dagger$} & 12.27 & 27.86 & 0.6927 \\ \hline
\multicolumn{1}{l|}{\our-f16f8} & 15.38 & 19.32 & 0.6607 \\
\multicolumn{1}{l|}{\our-f16f8-G} & \textbf{11.24} & 26.82 & \textbf{0.7046} \\ \hline
\multicolumn{4}{c}{Methods with external pre-trained CLIP} \\ \hline
\multicolumn{1}{l|}{LAFITE-CLIP$^\ddag$~\cite{zhou2021lafite} } & \textbf{8.12} & \textbf{32.24} & 0.7915 \\ \hline
\multicolumn{1}{l|}{\our-f16f8-G-CLIPr} & 8.97 & 27.43 & \textbf{0.7991}
\end{tabular}}
\vspace{-3mm}
\caption{Text-to-image generation on COCO.
For LDM scores, $T=250$; for \our, $T=200$.
$^\dagger$: reproduced with official code and configs.
$^\ddag$: obtained from official model checkpoints. G: classifier-free guidance with $\text{scale}=2.0$.
Note that LAFITE used CLIP at training, while \our uses it at inference only (CLIPr).}
\label{exp:T2I}
\vspace{-4mm}
\end{table}

\section{Experiments}
\label{sec:exp}
In this section, we empirically demonstrate that \our generates high-quality complex scene images that are also consistent to the multi-modal conditions, through the lens of text-to-image, scene-graph-to-image, and label-to-image generation tasks.
Moreover, to emphasize the capability of capturing multiple objects in the images globally, we conduct experiments on layout-to-image generation.
Lastly, extensive analyses are performed to validate design choices. We show that \our achieved state-of-the-art FID scores on multiple tasks under 5 settings with improved inference speed.

\paragraph{Notations} \our can be trained with different feature resolutions and levels. For simplicity and readability, a latent feature map where each feature corresponding to $n \cross n$ original image pixels is denoted f$n$. For example, a \our to generate 256$\cross$256 images using 32$\cross$32 high-level and 64$\cross$64 low-level latent code is denoted \our-f8f4. For LDM baselines, LDM-$n$ encodes $n \cross n$ pixels per feature.

\subsection{Datasets and Evaluation}
The main tasks we considered are text-to-image generation~(T2I) on COCO~\cite{lin2014microsoft}, scene-graph-to-image generation~(SG2I) on COCO-stuff and Visual Genome~(VG)~\cite{krishna2017visual}, label-to-image generation~(Label2I)~\cite{jyothi2019layoutvae} on COCO-stuff~\cite{lin2014microsoft}, and layout-to-image generation~(Layout2I) on COCO-stuff and OpenImages~\cite{kuznetsova2020open}. The standard metrics used to evaluate image synthesis tasks are Fréchet inception distance~(FID)~\cite{heusel2017gans} and Inception score~(IS)~\cite{salimans2016improved}. In addition, we considered other task-specific metrics such as CLIP score~\cite{hessel2021clipscore}, Precision and Recall~\cite{sajjadi2018assessing}, SceneFID~\cite{sylvain2021object}, YOLO score~\cite{li2021image}, PSNR~\cite{hore2010image}, and SSIM~\cite{wang2004image} when applicable. Please see the supplementary for detailed settings. For completeness, we also conducted user preference studies and experimented on unconditional image generation (UIG), including LSUN-bed~\cite{yu2015lsun}, CelebA-HQ~\cite{CelebAMask-HQ}, and Lanscape~\cite{skorokhodov2021aligning}. Due to the page limit, please refer to the supplementary for more results.

% \footnote{\url{https://arxiv.org/abs/2208.13753}}
\subsection{Conditional Complex Scene Generation}
\label{sec:cond}

\subsubsection{Text Conditional Image Generation}
\label{sec:cond}
We first experiment on the standard text-to-image (T2I) generation for COCO, and the results are shown in Table~\ref{exp:T2I}. We consider standard setting of training on COCO train2014 split. Orthogonal to recent T2I models pre-trained on huge image-text pairs, our goal is to synthesize images from diverse conditions.
In this setting, FID measures the image quality and CLIP-Score assesses the image-text consistency.
For completeness, IS is also reported, though FID is known for a stronger correlation with human judgment than IS~\cite{zhang2021cross,  sylvain2021object}.
Besides standard diffusion inference, we also report the variant with classifier-free guidance~\cite{nichol2021glide}.
As shown in Table~\ref{exp:T2I}, for both inference types, \our significantly outperforms the previous best model LDM by $\approx 2$ points for FID and $\approx 1$ point for CLIP-Score, achieving state-of-the-art scores on FID (15.38 vs. 11.24) and CLIP-Score (0.6607 vs. 0.7046).
In a different setting, LAFITE~\cite{zhou2021lafite} incorporated pre-trained CLIP~\cite{radford2021learning}, which contained abundant text-image knowledge from web-scale data pairs.
As an initial step for incorporating CLIP knowledge with \our, we report the results with a test-time only CLIP ranking trick~\cite{ding2021cogview} (10 inferences).
We can see that CLIPr further improves all metrics significantly, achieving comparable FID and CLIP-Score to LAFITE. An orthogonal direction to utilize CLIP at training similar to LAFITE is left to future works.

\begin{table}[t]
\centering
\resizebox{0.45\textwidth}{!}{
\setlength{\tabcolsep}{1.2mm}{
\begin{tabular}{l|ccc|ccc}
\multirow{2}{*}{Methods} & \multicolumn{3}{c|}{COCO} & \multicolumn{3}{c}{Visual Genome} \\ \cline{2-7} 
 & FID$\downarrow$ & IS$\uparrow$ & CLIP$\uparrow$ & FID$\downarrow$ & IS$\uparrow$ & CLIP$\uparrow$ \\ \hline
GT & - & - & 0.766 & - & - & 0.662 \\
Sg2Im &  127.0 & 6.179 & - & - & - & - \\
WSGC & 119.1 & 7.235 & - & 45.7 & 10.69 & - \\
LDMs-8$^\dagger$ &  49.14 & 13.33 & 0.627 & 36.88 & 14.60  & 0.611 \\
\hline
\our-f16f8 & \textbf{46.11} & \textbf{13.41} & \textbf{0.642} & \textbf{31.61} & \textbf{15.07} &  \textbf{0.613}
\end{tabular}}}
\vspace{-2mm}
\caption{Scene-graph-to-image generation on COCO and Visual Genome.
$^\dagger$: reproduced with official code and configs. Note that both LDM and \our are inferenced with classifier-free guidance.}
\vspace{-3mm}
\label{exp:sg2im}
\end{table}

\subsubsection{Image Generation from Scene Graph}
To further verify the claimed semantic relation capturing, we run SG2I on COCO-stuff and VG datasets, and the results are shown in Table~\ref{exp:sg2im}.
Clearly, \our outperforms all previous methods, including sg2im~\cite{johnson2018image}, WSGC~\cite{herzig2020learning}, and LDMs, in terms of FID and IS, achieving new state-of-the-art.
Moreover, to quantitatively measure the semantic correctness of the image w.r.t. its SG condition, we transform the SG to captions by concatenating the relation triplets (i.e., subject-predicate-object) and report the CLIP-score of the resulting image-caption pairs.
Our model surpasses previous work by $\approx 2\%$ on COCO and $\approx 0.2\%$ on VG.
This empirically verifies that, with the feature pyramid and coarse-to-fine generation strategy, \our improves modeling of complex relations.

\begin{table}[t]
\centering
\setlength{\tabcolsep}{1.5mm}{ 
\begin{tabular}{lcccc}
\multicolumn{1}{l|}{Name} & FID & IS & Precision & Recall \\ \hline
\multicolumn{5}{c}{3-8 labels in the image} \\ \hline
\multicolumn{1}{l|}{LayoutVAE} & 60.7 & - & - & - \\
\multicolumn{1}{l|}{+LostGAN} & 74.06 & 11.66 & 0.231 & 0.473 \\ 
\multicolumn{1}{l|}{LDMs-8$^\dagger$} & 51.45 & \textbf{15.05} & 0.434 & 0.576 \\
\multicolumn{1}{l|}{\our-f16f8} & \textbf{47.39} & 14.73 &\textbf{0.437} & \textbf{0.595} \\ \hline
\multicolumn{5}{c}{2-30 labels in the image} \\ \hline
\multicolumn{1}{l|}{LDMs-8$^\dagger$} & 29.17 & \textbf{18.00} & 0.563 & \textbf{0.554} \\
\multicolumn{1}{l|}{\our-f16f8} & \textbf{27.65} & 17.70 & \textbf{0.573} & 0.542
\end{tabular}}
\vspace{-2mm}
\caption{Label-to-image generation on COCO.
$^\dagger$: reproduced with official code and configs.
}
\vspace{-2mm}
\label{exp:label2I_coco}
\end{table}
\subsubsection{Label-to-Image Generation}
Label-to-image produces scene image conditioned on image-level labels.
Unlike T2I or SG2I, where scene structure is specified by the text conditions, this task requires a model to combine objects more freely and synthesize a coherent image.
In addition to FID and IS, precision and recall are reported for object-level quality and diversity measurement, respectively.
We conduct experiments on Label2I with COCO-stuff. As the shown in Table~\ref{exp:label2I_coco}, our model outperforms all previous approaches, including LayoutVAE~\cite{jyothi2019layoutvae}\footnote{LayoutVAE implements Label2I as Label2Layout + Layout2I and reported 128 resolution result. We follow~\citet{yang2021layouttransformer} to adopt LostGAN~\cite{sun2019image} to achieve 256 resolution.} and LDMs, on not only FID but also precision and recall under the more common 3-8 labels setting.
This indicates that, \our achieves better image quality and data manifold modeling of multi-object images.
We further challenge \our with a harder 2-30 labels setting and still establish SOTA FID.

\begin{table}[t]
\center
\resizebox{0.48\textwidth}{!}{
\setlength{\tabcolsep}{0.5mm}{
\begin{tabular}{l|ccc|cc}
\multirow{2}{*}{Methods} & \multicolumn{3}{c|}{COCO 256} & \multicolumn{2}{c}{OpenImage 256} \\ \cline{2-6} 
 & FID$\downarrow$ & YOLO$\uparrow$ & SceneFID$\downarrow$ & FID$\downarrow$ & SceneFID$\downarrow$ \\ \hline
LostGAN-V2 & 42.55 & - & - & - & - \\
OC-GAN & 41.65 & - & - & -  & -\\
SPADE & 41.11 & - & - & -  & - \\
VQGAN+T & 56.58 & - & 24.07 & 45.33 &  15.85 \\
LDM-8 (100 steps) & 42.06 & - & - & - & - \\
LDM-8$^\dagger$ (100 steps) & 41.02 & 14.67 & 21.63 & -  & - \\
LDM-4 (200 steps) & 40.91 & - & - & 32.02 & - \\ \hline

\our-f16f8 (100 steps) & 38.95 & 16.71 & 17.69 & -  & - \\ 
\our-f8f4 (200 steps) & \textbf{37.14} & \textbf{17.22 } & \textbf{14.91} & \textbf{29.04}  & \textbf{12.77} \\
\end{tabular}}}
\vspace{-2mm}
\caption{Layout-to-image generation on COCO (segmentation challenge split) and OpenImages. $^\dagger$: reproduced with official code and configs.}
\vspace{-4mm}
\label{exp:layout2i}
\end{table}

% \begin{table}[t]
% \center
% \resizebox{0.48\textwidth}{!}{
% \setlength{\tabcolsep}{0.5mm}{
% \begin{tabular}{l|ccc|cc}
% \multirow{2}{*}{Methods} & \multicolumn{3}{c|}{COCO 256} & \multicolumn{2}{c}{OpenImage 256} \\ \cline{2-6} 
%  & FID$\downarrow$ & IS$\uparrow$ & SceneFID$\downarrow$ & FID$\downarrow$ & SceneFID$\downarrow$ \\ \hline
% LostGAN-V2 & 42.55 & - & - & - & - \\
% OC-GAN & 41.65 & - & - & -  & -\\
% SPADE & 41.11 & - & - & -  & - \\
% VQGAN+T & 56.58 & - & 24.07 & 45.33 &  15.85 \\
% LDM-8 (100 steps) & 42.06 & - & - & - & - \\
% LDM-8$^\dagger$ (100 steps) & 41.02 & 17.01 & 21.63 & -  & - \\
% LDM-4 (200 steps) & 40.91 & - & - & 32.02 & - \\ \hline

% \our-f16f8 (100 steps) & 38.95 & 16.57 & 17.69 & -  & - \\ 
% \our-f8f4 (200 steps) & \textbf{37.14} & \textbf{18.62} & \textbf{14.91} & \textbf{29.04}  & \textbf{12.77} \\
% \end{tabular}}}
% % \vspace{-2mm}
% \caption{Layout-to-image generation on COCO and OpenImages. $^\dagger$: reproduced with official code and configs.}
% % \vspace{-3mm}
% \label{exp:layout2i}
% \end{table}

\subsubsection{Layout-to-Image Generation}
Our Layout2I results show-cases that multiple objects' shapes and details can be synthesized. Specifically, we compare our \our with previous methods under two different settings. Firstly, we follow LDM and perform experiments on COCO stuff segmentation challenge split and OpenImage datasets.
The results are shown in Table~\ref{exp:layout2i}. One can find that \our outperforms previous methods, including LostGAN-v2~\cite{sun2019image}, OC-GAN~\cite{sylvain2021object}, SPADE~\cite{SPADE2019}, VQGAN+T (combining \citealt{esser2021taming} and \citealt{brown2020language}), and LDM, on FID by at least $2$ points, achieving new state-of-the-art for both COCO and OpenImages.
Moreover, we achieve the best YOLO scores and sceneFID, indicating the most visually realistic instance-level objects. Secondly, we follow TwFA~\cite{yang2022modeling} and conduct experiments on standard COCO stuff and Visual Genome datasets. Please refer to the supplementary material for more detail.

\begin{figure}[t]
  \centering
  \includegraphics[page=10,trim={190 45 175 55}, clip, width=0.45\textwidth]{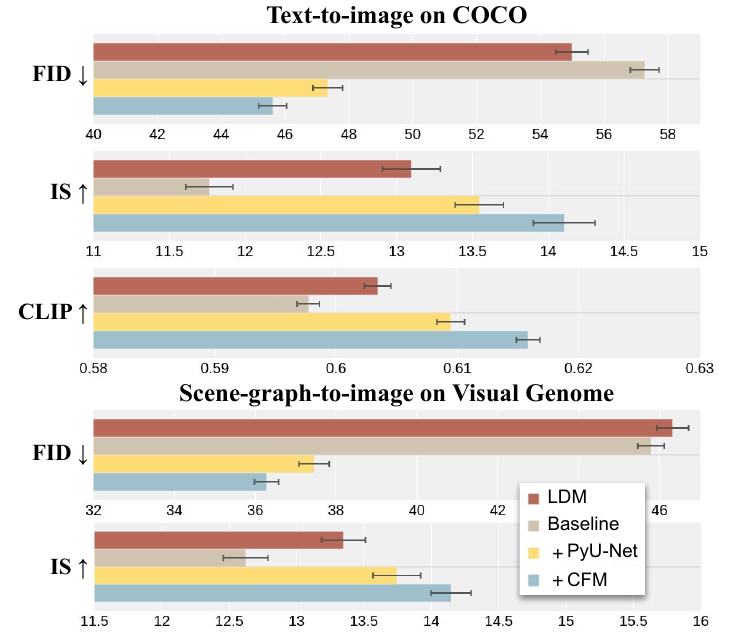}
  \vspace{-2mm}
  \caption{Model ablation on COCO T2I and VG SG2I. CFM denotes our coarse-to-fine modulation.}
  \vspace{-3mm}
  \label{exp:model_ablation}
\end{figure}

\subsection{Model Analysis}
\label{sec:model_ana}

\subsubsection{Model Ablation}

To verify the key novel designs of \our, we perform ablation studies on two tasks: text-to-image (T2I) on COCO and scene-graph-to-image (SG2I) on Visual Genome.
Figure~\ref{exp:model_ablation} showcases the contribution of each deployed component in \our. For ablations and hyper-parameter tunings, we train for 250K iterations to allow more experiments. Models with the best dev scores are further trained to obtain the final test scores in Sec.~\ref{sec:cond}.
We report mean and the corresponding 95\% confidence interval by conducting the bootstrap test~\cite{koehn2004statistical} and the sample size equals to test set size; resampled for 100 times.
For the baseline, we use two LDMs and perform a simple sequential learning strategy.
More specifically, the first LDM learns the distribution of the high-level feature map (LDM-16); and the second LDM is deployed to model the low-level feature of f8 (LDM-8).
In this baseline model, we concat LDM-8's output feature map and the denoising target feature feed into the LDM-8 for denoising.
To justify the shared U-Net design of \ourunet, we first apply this module without CFM to the baseline.
Sharing U-Net reduces the model parameters from 1.18B (baseline) to 590M (baseline + \ourunet).
Finally, the coarse-to-fine modulation is added, with only a minor increase in parameter count (total of 697M), and performance is further boosted for all metrics.
We can see that each component significantly improves the generation results; models with \ourunet and CFM are significantly better than the LDMs on all metrics.

\begin{figure}[t]
  \centering
  \includegraphics[page=3
  ,trim={140 95 145 95}, clip, width=0.42\textwidth]{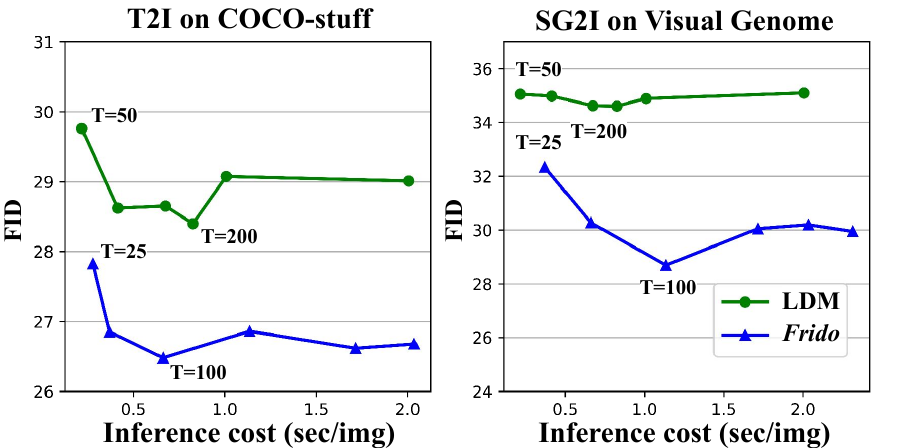}
  \vspace{-1mm}
  \caption{Speed-quality trade-off analysis. Lower FID indicates better image quality.}
  \vspace{-2mm}
  \label{exp:inference_cost}
\end{figure}

\subsubsection{Computation Cost Analysis}
Here we analyze the inference cost of our model.
In Fig.~\ref{exp:inference_cost}, we compare \our with LDM on the speed-quality trade-off. 
In this figure, we inference each model with different inference timesteps $T$ and then plot the FID scores and against the per-image inference cost. Note that the experiments are done on validation splits with batch size of 32 using 1 V100.
It is demonstrated that under similar inference budget, \our achieve decent performance gain comparing to LDM, confirming the claimed efficiency of our model.
For other comparisons on FLOPs, parameter counts, and inference time, please see the supplementary.
Note that by operating in the latent space, LDM is among the faster ones within the DM model class.
\our further reduce the cost by putting part of the denoising load at lower-resolution.

\paragraph{Takeaways}
The empirical studies have shown that \our significantly outperforms the baseline LDM for complex scene image synthesis, and even achieves SOTA in 5 settings.
Our modeling novelty, including \ourunet and the coarse-to-fine modulation, are statistically effective.
Last but not least, \our is more efficient, as can be seen in a head-to-head comparison to LDM, which mitigates the notorious heavy inference cost for diffusion models.

\section{Related Work}
\label{sec:related}

\paragraph{More Generative Models for Image Synthesis}
The community has witnessed great progress of image synthesis in the past decade.
Other than the previously discussed works, the families of GANs~\cite{hu2021text, xu2018attngan, karras2019style, brock2018large, gafni2022make, zhang2021cross, hinz2020semantic, karras2021alias}, VAEs~\cite{sohn2015learning}, autoregressive models~\cite{razavi2019generating, chang2022maskgit, yu2021vector}, flow-based methods~\cite{dinh2016density}, and diffusion-based models~\cite{saharia2022photorealistic, gu2022vector} have all made great contribution to shape this field.
\our is a hybrid of the VAE and DM family, combining the best of both worlds for outstanding image quality on complex scenes, and significantly improved DM inference.
Very recently, large-scale pre-training for text-to-image generation~\cite{ramesh2022hierarchical} has gained vast attention and achieved superior results.
\our is orthogonal to these models, as we investigate coarse-to-fine synthesis and multi-modality inputs beyond text.

\paragraph{Two-Stage Generative Models}
Recently, many two-stage generative models~\cite{van2016pixel, jahn2021high, pandey2022diffusevae, zhou2021lafite} are proposed to tackle the drawbacks of the one-stage models.
The representative VQ-VAE~\cite{van2017neural} first encodes an image into a discrete latent space with a lower spatial resolution and then uses an auto-regressive network to model such space.
The first step is called Vector Quantization (VQ), which reduces the input information to allow auto-encoder learning.
In addition, VQ bridges images to other modalities, such as language~\cite{ding2021cogview, chen2020generative} and audio ~\cite{yan2021videogpt}, seamlessly by converting to discrete tokens.
In the second stage, an auto-regressive(e.g. PixelCNN~\cite{van2016conditional}, VQGAN) or diffusion model (LDM, VQ-Diffusion~\cite{tang2022improved}) is adopted to model the encoded latent space.
\our contributes to both stages by proposing MS-VQGAN and \ourunet for DM.

\paragraph{Coarse-to-Fine Image Generation Approaches}
Instead of generating a full-resolution image in one step, coarse-to-fine generation synthesizes an image with multiple steps, from low to high resolution in pixel space~\cite{gregor2015draw, mansimov2015generating, ho2022cascaded} or from high-level to low-level information in latent space~\cite{razavi2019generating, child2019generating}.
These allow model to better capture the information in different levels and have shown to achieve higher quality.
For instance, AttnGAN~\cite{xu2018attngan} and StackGAN~\cite{zhang2017stackgan, zhang2018stackgan++} first produce an image in low resolution (e.g., 1/8 of the full-size) and then iteratively scale up the generated image until achieving the final resolution.
Different from the above works, we share the core network for each scale. Therefore, the overhead compared to single-scale models is minimized.

\section{Conclusion}
\label{sec:conclusion}
We propose \our, a new image generative model, empowering an under-explored coarse-to-fine prior in the diffusion model family.
Key designs such as multi-scale codebooks, a single shared U-Net, and the special modulation mechanism are shown to be effective via extensive experiments.
Empirically, we apply this model to a diverse set of cross-modal image synthesis tasks and achieve 5 new state-of-the-art results.
From a practical aspect, \our also mitigates the well-known slow inference pain-point of diffusion methods.

\clearpage

\section{Acknowledgement}
This work is supported in part by the National Science and Technology Council of
Taiwan under grant NSTC 111-2634-F-002-020. We also thank the National Center for High-performance Computing (NCHC), Taiwan Computing Cloud (TWCC), and Novatek Foundation for providing computational resources and extra funding.
Moreover, we are immensely grateful to Andreas Blattmann, a co-author of LDM~\cite{rombach2022high}, for sharing his valuable experience in implementing diffusion models, which was very helpful in the early stage of this work.
Last but not least, we thank Cheng-Fu Yang and Chiao-An Yang, our colleagues in the Vision and Learning Lab at NTU, for their feedback that helps polish this paper.

\bibliography{reference} 

\begin{thebibliography}{90}
\providecommand{\natexlab}[1]{#1}

\bibitem[{Anokhin et~al.(2021)Anokhin, Demochkin, Khakhulin, Sterkin,
  Lempitsky, and Korzhenkov}]{anokhin2021image}
Anokhin, I.; Demochkin, K.; Khakhulin, T.; Sterkin, G.; Lempitsky, V.; and
  Korzhenkov, D. 2021.
\newblock Image generators with conditionally-independent pixel synthesis.
\newblock In \emph{CVPR}.

\bibitem[{Bochkovskiy, Wang, and Liao(2020)}]{bochkovskiy2020yolov4}
Bochkovskiy, A.; Wang, C.-Y.; and Liao, H.-Y.~M. 2020.
\newblock Yolov4: Optimal speed and accuracy of object detection.
\newblock \emph{arXiv preprint arXiv:2004.10934}.

\bibitem[{Brock, Donahue, and Simonyan(2019)}]{brock2018large}
Brock, A.; Donahue, J.; and Simonyan, K. 2019.
\newblock Large scale GAN training for high fidelity natural image synthesis.
\newblock In \emph{ICLR}.

\bibitem[{Brown et~al.(2020)Brown, Mann, Ryder, Subbiah, Kaplan, Dhariwal,
  Neelakantan, Shyam, Sastry, Askell et~al.}]{brown2020language}
Brown, T.; Mann, B.; Ryder, N.; Subbiah, M.; Kaplan, J.~D.; Dhariwal, P.;
  Neelakantan, A.; Shyam, P.; Sastry, G.; Askell, A.; et~al. 2020.
\newblock Language models are few-shot learners.
\newblock In \emph{NeurIPS}.

\bibitem[{Brundage et~al.(2018)Brundage, Avin, Clark, Toner, Eckersley,
  Garfinkel, Dafoe, Scharre, Zeitzoff, Filar et~al.}]{brundage2018malicious}
Brundage, M.; Avin, S.; Clark, J.; Toner, H.; Eckersley, P.; Garfinkel, B.;
  Dafoe, A.; Scharre, P.; Zeitzoff, T.; Filar, B.; et~al. 2018.
\newblock The malicious use of artificial intelligence: Forecasting,
  prevention, and mitigation.
\newblock \emph{arXiv preprint arXiv:1802.07228}.

\bibitem[{Casanova et~al.(2021)Casanova, Careil, Verbeek, Drozdzal, and
  Romero~Soriano}]{casanova2021instance}
Casanova, A.; Careil, M.; Verbeek, J.; Drozdzal, M.; and Romero~Soriano, A.
  2021.
\newblock Instance-conditioned gan.
\newblock In \emph{NeurIPS}.

\bibitem[{Chang et~al.(2022)Chang, Zhang, Jiang, Liu, and
  Freeman}]{chang2022maskgit}
Chang, H.; Zhang, H.; Jiang, L.; Liu, C.; and Freeman, W.~T. 2022.
\newblock Maskgit: Masked generative image transformer.
\newblock In \emph{CVPR}.

\bibitem[{Chen et~al.(2020)Chen, Radford, Child, Wu, Jun, Luan, and
  Sutskever}]{chen2020generative}
Chen, M.; Radford, A.; Child, R.; Wu, J.; Jun, H.; Luan, D.; and Sutskever, I.
  2020.
\newblock Generative pretraining from pixels.
\newblock In \emph{ICML}.

\bibitem[{Child et~al.(2019)Child, Gray, Radford, and
  Sutskever}]{child2019generating}
Child, R.; Gray, S.; Radford, A.; and Sutskever, I. 2019.
\newblock Generating long sequences with sparse transformers.
\newblock \emph{arXiv preprint arXiv:1904.10509}.

\bibitem[{Devlin et~al.(2019)Devlin, Chang, Lee, and
  Toutanova}]{devlin2018bert}
Devlin, J.; Chang, M.-W.; Lee, K.; and Toutanova, K. 2019.
\newblock Bert: Pre-training of deep bidirectional transformers for language
  understanding.
\newblock In \emph{NAACL-HLT}.

\bibitem[{Dhariwal and Nichol(2021)}]{dhariwal2021diffusion}
Dhariwal, P.; and Nichol, A. 2021.
\newblock Diffusion models beat gans on image synthesis.
\newblock In \emph{NeurIPS}.

\bibitem[{Ding et~al.(2021)Ding, Yang, Hong, Zheng, Zhou, Yin, Lin, Zou, Shao,
  Yang et~al.}]{ding2021cogview}
Ding, M.; Yang, Z.; Hong, W.; Zheng, W.; Zhou, C.; Yin, D.; Lin, J.; Zou, X.;
  Shao, Z.; Yang, H.; et~al. 2021.
\newblock Cogview: Mastering text-to-image generation via transformers.
\newblock In \emph{NeurIPS}.

\bibitem[{Dinh, Krueger, and Bengio(2014)}]{dinh2014nice}
Dinh, L.; Krueger, D.; and Bengio, Y. 2014.
\newblock Nice: Non-linear independent components estimation.
\newblock \emph{arXiv preprint arXiv:1410.8516}.

\bibitem[{Dinh, Sohl-Dickstein, and Bengio(2017)}]{dinh2016density}
Dinh, L.; Sohl-Dickstein, J.; and Bengio, S. 2017.
\newblock Density estimation using real nvp.
\newblock In \emph{ICLR}.

\bibitem[{Elfwing, Uchibe, and Doya(2018)}]{elfwing2018sigmoid}
Elfwing, S.; Uchibe, E.; and Doya, K. 2018.
\newblock Sigmoid-weighted linear units for neural network function
  approximation in reinforcement learning.
\newblock \emph{Neural Networks}.

\bibitem[{Esser et~al.(2021)Esser, Rombach, Blattmann, and
  Ommer}]{esser2021imagebart}
Esser, P.; Rombach, R.; Blattmann, A.; and Ommer, B. 2021.
\newblock Imagebart: Bidirectional context with multinomial diffusion for
  autoregressive image synthesis.
\newblock In \emph{NeurIPS}.

\bibitem[{Esser, Rombach, and Ommer(2021)}]{esser2021taming}
Esser, P.; Rombach, R.; and Ommer, B. 2021.
\newblock Taming transformers for high-resolution image synthesis.
\newblock In \emph{CVPR}.

\bibitem[{Falcon and team(2019)}]{pytorchLightning}
Falcon, W.; and team, T. P.~L. 2019.
\newblock PyTorch Lightning.
\newblock \emph{https://www.pytorchlightning.ai}.

\bibitem[{Gafni et~al.(2022)Gafni, Polyak, Ashual, Sheynin, Parikh, and
  Taigman}]{gafni2022make}
Gafni, O.; Polyak, A.; Ashual, O.; Sheynin, S.; Parikh, D.; and Taigman, Y.
  2022.
\newblock Make-a-scene: Scene-based text-to-image generation with human priors.
\newblock In \emph{ECCV}.

\bibitem[{Goodfellow et~al.(2014)Goodfellow, Pouget-Abadie, Mirza, Xu,
  Warde-Farley, Ozair, Courville, and Bengio}]{goodfellow2014generative}
Goodfellow, I.; Pouget-Abadie, J.; Mirza, M.; Xu, B.; Warde-Farley, D.; Ozair,
  S.; Courville, A.; and Bengio, Y. 2014.
\newblock Generative adversarial nets.
\newblock In \emph{NeurIPS}.

\bibitem[{Gregor et~al.(2015)Gregor, Danihelka, Graves, Rezende, and
  Wierstra}]{gregor2015draw}
Gregor, K.; Danihelka, I.; Graves, A.; Rezende, D.; and Wierstra, D. 2015.
\newblock Draw: A recurrent neural network for image generation.
\newblock In \emph{ICML}.

\bibitem[{Gu et~al.(2022)Gu, Chen, Bao, Wen, Zhang, Chen, Yuan, and
  Guo}]{gu2022vector}
Gu, S.; Chen, D.; Bao, J.; Wen, F.; Zhang, B.; Chen, D.; Yuan, L.; and Guo, B.
  2022.
\newblock Vector quantized diffusion model for text-to-image synthesis.
\newblock In \emph{CVPR}.

\bibitem[{He et~al.(2021)He, Liao, Yang, Yang, Song, Rosenhahn, and
  Xiang}]{he2021context}
He, S.; Liao, W.; Yang, M.~Y.; Yang, Y.; Song, Y.-Z.; Rosenhahn, B.; and Xiang,
  T. 2021.
\newblock Context-aware layout to image generation with enhanced object
  appearance.
\newblock In \emph{CVPR}.

\bibitem[{Herzig et~al.(2020)Herzig, Bar, Xu, Chechik, Darrell, and
  Globerson}]{herzig2020learning}
Herzig, R.; Bar, A.; Xu, H.; Chechik, G.; Darrell, T.; and Globerson, A. 2020.
\newblock Learning canonical representations for scene graph to image
  generation.
\newblock In \emph{ECCV}.

\bibitem[{Hessel et~al.(2021)Hessel, Holtzman, Forbes, Bras, and
  Choi}]{hessel2021clipscore}
Hessel, J.; Holtzman, A.; Forbes, M.; Bras, R.~L.; and Choi, Y. 2021.
\newblock Clipscore: A reference-free evaluation metric for image captioning.
\newblock In \emph{EMNLP}.

\bibitem[{Heusel et~al.(2017)Heusel, Ramsauer, Unterthiner, Nessler, and
  Hochreiter}]{heusel2017gans}
Heusel, M.; Ramsauer, H.; Unterthiner, T.; Nessler, B.; and Hochreiter, S.
  2017.
\newblock Gans trained by a two time-scale update rule converge to a local nash
  equilibrium.
\newblock In \emph{NeurIPS}.

\bibitem[{Hinz, Heinrich, and Wermter(2020)}]{hinz2020semantic}
Hinz, T.; Heinrich, S.; and Wermter, S. 2020.
\newblock Semantic object accuracy for generative text-to-image synthesis.
\newblock \emph{TPAMI}.

\bibitem[{Ho, Jain, and Abbeel(2020)}]{ho2020denoising}
Ho, J.; Jain, A.; and Abbeel, P. 2020.
\newblock Denoising diffusion probabilistic models.
\newblock In \emph{NeurIPS}.

\bibitem[{Ho et~al.(2022)Ho, Saharia, Chan, Fleet, Norouzi, and
  Salimans}]{ho2022cascaded}
Ho, J.; Saharia, C.; Chan, W.; Fleet, D.~J.; Norouzi, M.; and Salimans, T.
  2022.
\newblock Cascaded Diffusion Models for High Fidelity Image Generation.
\newblock \emph{JMLR}.

\bibitem[{Hore and Ziou(2010)}]{hore2010image}
Hore, A.; and Ziou, D. 2010.
\newblock Image quality metrics: PSNR vs. SSIM.
\newblock In \emph{ICPR}.

\bibitem[{Huang and Belongie(2017)}]{huang2017arbitrary}
Huang, X.; and Belongie, S. 2017.
\newblock Arbitrary style transfer in real-time with adaptive instance
  normalization.
\newblock In \emph{ICCV}.

\bibitem[{Jahn, Rombach, and Ommer(2021)}]{jahn2021high}
Jahn, M.; Rombach, R.; and Ommer, B. 2021.
\newblock High-Resolution Complex Scene Synthesis with Transformers.
\newblock \emph{arXiv preprint arXiv:2105.06458}.

\bibitem[{Johnson, Gupta, and Li(2018)}]{johnson2018image}
Johnson, J.; Gupta, A.; and Li, F.-F. 2018.
\newblock Image generation from scene graphs.
\newblock In \emph{CVPR}.

\bibitem[{Jyothi et~al.(2019)Jyothi, Durand, He, Sigal, and
  Mori}]{jyothi2019layoutvae}
Jyothi, A.~A.; Durand, T.; He, J.; Sigal, L.; and Mori, G. 2019.
\newblock Layoutvae: Stochastic scene layout generation from a label set.
\newblock In \emph{ICCV}.

\bibitem[{Karras et~al.(2017)Karras, Aila, Laine, and
  Lehtinen}]{karras2017progressive}
Karras, T.; Aila, T.; Laine, S.; and Lehtinen, J. 2017.
\newblock Progressive growing of gans for improved quality, stability, and
  variation.
\newblock \emph{arXiv preprint arXiv:1710.10196}.

\bibitem[{Karras et~al.(2021)Karras, Aittala, Laine, H{\"a}rk{\"o}nen,
  Hellsten, Lehtinen, and Aila}]{karras2021alias}
Karras, T.; Aittala, M.; Laine, S.; H{\"a}rk{\"o}nen, E.; Hellsten, J.;
  Lehtinen, J.; and Aila, T. 2021.
\newblock Alias-free generative adversarial networks.
\newblock In \emph{NeurIPS}.

\bibitem[{Karras, Laine, and Aila(2019)}]{karras2019style}
Karras, T.; Laine, S.; and Aila, T. 2019.
\newblock A style-based generator architecture for generative adversarial
  networks.
\newblock In \emph{CVPR}.

\bibitem[{Karras et~al.(2020)Karras, Laine, Aittala, Hellsten, Lehtinen, and
  Aila}]{karras2020analyzing}
Karras, T.; Laine, S.; Aittala, M.; Hellsten, J.; Lehtinen, J.; and Aila, T.
  2020.
\newblock Analyzing and improving the image quality of stylegan.
\newblock In \emph{CVPR}.

\bibitem[{Kim et~al.(2022)Kim, Shin, Song, Kang, and Moon}]{kim2022soft}
Kim, D.; Shin, S.; Song, K.; Kang, W.; and Moon, I.-C. 2022.
\newblock Soft truncation: A universal training technique of score-based
  diffusion model for high precision score estimation.
\newblock In \emph{ICML}.

\bibitem[{Kingma and Dhariwal(2018)}]{kingma2018glow}
Kingma, D.~P.; and Dhariwal, P. 2018.
\newblock Glow: Generative flow with invertible 1x1 convolutions.
\newblock In \emph{NeurIPS}.

\bibitem[{Kingma and Welling(2014)}]{kingma2013auto}
Kingma, D.~P.; and Welling, M. 2014.
\newblock Auto-encoding variational bayes.
\newblock In \emph{ICLR}.

\bibitem[{Koehn(2004)}]{koehn2004statistical}
Koehn, P. 2004.
\newblock Statistical significance tests for machine translation evaluation.
\newblock In \emph{EMNLP}.

\bibitem[{Krishna et~al.(2017)Krishna, Zhu, Groth, Johnson, Hata, Kravitz,
  Chen, Kalantidis, Li, Shamma et~al.}]{krishna2017visual}
Krishna, R.; Zhu, Y.; Groth, O.; Johnson, J.; Hata, K.; Kravitz, J.; Chen, S.;
  Kalantidis, Y.; Li, L.-J.; Shamma, D.~A.; et~al. 2017.
\newblock Visual genome: Connecting language and vision using crowdsourced
  dense image annotations.
\newblock \emph{IJCV}.

\bibitem[{Kuznetsova et~al.(2020)Kuznetsova, Rom, Alldrin, Uijlings, Krasin,
  Pont-Tuset, Kamali, Popov, Malloci, Kolesnikov et~al.}]{kuznetsova2020open}
Kuznetsova, A.; Rom, H.; Alldrin, N.; Uijlings, J.; Krasin, I.; Pont-Tuset, J.;
  Kamali, S.; Popov, S.; Malloci, M.; Kolesnikov, A.; et~al. 2020.
\newblock The open images dataset v4.
\newblock \emph{IJCV}.

\bibitem[{Kynk{\"a}{\"a}nniemi et~al.(2019)Kynk{\"a}{\"a}nniemi, Karras, Laine,
  Lehtinen, and Aila}]{kynkaanniemi2019improved}
Kynk{\"a}{\"a}nniemi, T.; Karras, T.; Laine, S.; Lehtinen, J.; and Aila, T.
  2019.
\newblock Improved precision and recall metric for assessing generative models.
\newblock In \emph{NeurIPS}.

\bibitem[{Lee et~al.(2020)Lee, Liu, Wu, and Luo}]{CelebAMask-HQ}
Lee, C.-H.; Liu, Z.; Wu, L.; and Luo, P. 2020.
\newblock MaskGAN: Towards Diverse and Interactive Facial Image Manipulation.
\newblock In \emph{CVPR}.

\bibitem[{Li et~al.(2019)Li, Zhang, Zhang, Huang, He, Lyu, and
  Gao}]{li2019object}
Li, W.; Zhang, P.; Zhang, L.; Huang, Q.; He, X.; Lyu, S.; and Gao, J. 2019.
\newblock Object-driven text-to-image synthesis via adversarial training.
\newblock In \emph{CVPR}.

\bibitem[{Li et~al.(2021)Li, Wu, Koh, Tang, and Sun}]{li2021image}
Li, Z.; Wu, J.; Koh, I.; Tang, Y.; and Sun, L. 2021.
\newblock Image Synthesis from Layout with Locality-Aware Mask Adaption.
\newblock In \emph{ICCV}.

\bibitem[{Liao et~al.(2022)Liao, Hu, Yang, and Rosenhahn}]{hu2021text}
Liao, W.; Hu, K.; Yang, M.~Y.; and Rosenhahn, B. 2022.
\newblock Text to image generation with semantic-spatial aware GAN.
\newblock In \emph{CVPR}.

\bibitem[{Lin et~al.(2014)Lin, Maire, Belongie, Hays, Perona, Ramanan,
  Doll{\'a}r, and Zitnick}]{lin2014microsoft}
Lin, T.-Y.; Maire, M.; Belongie, S.; Hays, J.; Perona, P.; Ramanan, D.;
  Doll{\'a}r, P.; and Zitnick, C.~L. 2014.
\newblock Microsoft coco: Common objects in context.
\newblock In \emph{ECCV}.

\bibitem[{Mansimov et~al.(2016)Mansimov, Parisotto, Ba, and
  Salakhutdinov}]{mansimov2015generating}
Mansimov, E.; Parisotto, E.; Ba, J.~L.; and Salakhutdinov, R. 2016.
\newblock Generating images from captions with attention.
\newblock In \emph{ICLR}.

\bibitem[{Nichol et~al.(2022)Nichol, Dhariwal, Ramesh, Shyam, Mishkin, McGrew,
  Sutskever, and Chen}]{nichol2021glide}
Nichol, A.; Dhariwal, P.; Ramesh, A.; Shyam, P.; Mishkin, P.; McGrew, B.;
  Sutskever, I.; and Chen, M. 2022.
\newblock Glide: Towards photorealistic image generation and editing with
  text-guided diffusion models.
\newblock In \emph{ICML}.

\bibitem[{Nichol and Dhariwal(2021)}]{nichol2021improved}
Nichol, A.~Q.; and Dhariwal, P. 2021.
\newblock Improved denoising diffusion probabilistic models.
\newblock In \emph{ICML}.

\bibitem[{Pandey et~al.(2022)Pandey, Mukherjee, Rai, and
  Kumar}]{pandey2022diffusevae}
Pandey, K.; Mukherjee, A.; Rai, P.; and Kumar, A. 2022.
\newblock DiffuseVAE: Efficient, Controllable and High-Fidelity Generation from
  Low-Dimensional Latents.
\newblock \emph{TMLR}.

\bibitem[{Park et~al.(2019)Park, Liu, Wang, and Zhu}]{SPADE2019}
Park, T.; Liu, M.-Y.; Wang, T.-C.; and Zhu, J.-Y. 2019.
\newblock Semantic image synthesis with spatially-adaptive normalization.
\newblock In \emph{CVPR}.

\bibitem[{Parmar et~al.(2021)Parmar, Li, Lee, and Tu}]{parmar2021dual}
Parmar, G.; Li, D.; Lee, K.; and Tu, Z. 2021.
\newblock Dual contradistinctive generative autoencoder.
\newblock In \emph{CVPR}.

\bibitem[{Radford et~al.(2021)Radford, Kim, Hallacy, Ramesh, Goh, Agarwal,
  Sastry, Askell, Mishkin, Clark et~al.}]{radford2021learning}
Radford, A.; Kim, J.~W.; Hallacy, C.; Ramesh, A.; Goh, G.; Agarwal, S.; Sastry,
  G.; Askell, A.; Mishkin, P.; Clark, J.; et~al. 2021.
\newblock Learning transferable visual models from natural language
  supervision.
\newblock In \emph{ICML}.

\bibitem[{Radford, Metz, and Chintala(2015)}]{radford2015unsupervised}
Radford, A.; Metz, L.; and Chintala, S. 2015.
\newblock Unsupervised representation learning with deep convolutional
  generative adversarial networks.
\newblock \emph{arXiv preprint arXiv:1511.06434}.

\bibitem[{Ramesh et~al.(2022)Ramesh, Dhariwal, Nichol, Chu, and
  Chen}]{ramesh2022hierarchical}
Ramesh, A.; Dhariwal, P.; Nichol, A.; Chu, C.; and Chen, M. 2022.
\newblock Hierarchical text-conditional image generation with clip latents.
\newblock \emph{arXiv preprint arXiv:2204.06125}.

\bibitem[{Razavi, Van~den Oord, and Vinyals(2019)}]{razavi2019generating}
Razavi, A.; Van~den Oord, A.; and Vinyals, O. 2019.
\newblock Generating diverse high-fidelity images with vq-vae-2.
\newblock In \emph{NeurIPS}.

\bibitem[{Rombach et~al.(2022)Rombach, Blattmann, Lorenz, Esser, and
  Ommer}]{rombach2022high}
Rombach, R.; Blattmann, A.; Lorenz, D.; Esser, P.; and Ommer, B. 2022.
\newblock High-resolution image synthesis with latent diffusion models.
\newblock In \emph{CVPR}.

\bibitem[{Ronneberger, Fischer, and Brox(2015)}]{RonnebergerFB15}
Ronneberger, O.; Fischer, P.; and Brox, T. 2015.
\newblock U-net: Convolutional networks for biomedical image segmentation.
\newblock In \emph{MICCAI}.

\bibitem[{Saharia et~al.(2022)Saharia, Chan, Saxena, Li, Whang, Denton,
  Ghasemipour, Ayan, Mahdavi, Lopes et~al.}]{saharia2022photorealistic}
Saharia, C.; Chan, W.; Saxena, S.; Li, L.; Whang, J.; Denton, E.; Ghasemipour,
  S. K.~S.; Ayan, B.~K.; Mahdavi, S.~S.; Lopes, R.~G.; et~al. 2022.
\newblock Photorealistic Text-to-Image Diffusion Models with Deep Language
  Understanding.
\newblock In \emph{NeurIPS}.

\bibitem[{Sajjadi et~al.(2018)Sajjadi, Bachem, Lucic, Bousquet, and
  Gelly}]{sajjadi2018assessing}
Sajjadi, M.~S.; Bachem, O.; Lucic, M.; Bousquet, O.; and Gelly, S. 2018.
\newblock Assessing generative models via precision and recall.
\newblock In \emph{NeurIPS}.

\bibitem[{Salimans et~al.(2016)Salimans, Goodfellow, Zaremba, Cheung, Radford,
  and Chen}]{salimans2016improved}
Salimans, T.; Goodfellow, I.; Zaremba, W.; Cheung, V.; Radford, A.; and Chen,
  X. 2016.
\newblock Improved techniques for training gans.
\newblock In \emph{NeurIPS}.

\bibitem[{Sauer et~al.(2021)Sauer, Chitta, M{\"u}ller, and
  Geiger}]{sauer2021projected}
Sauer, A.; Chitta, K.; M{\"u}ller, J.; and Geiger, A. 2021.
\newblock Projected gans converge faster.
\newblock In \emph{NeurIPS}.

\bibitem[{Skorokhodov, Sotnikov, and Elhoseiny(2021)}]{skorokhodov2021aligning}
Skorokhodov, I.; Sotnikov, G.; and Elhoseiny, M. 2021.
\newblock Aligning latent and image spaces to connect the unconnectable.
\newblock In \emph{ICCV}.

\bibitem[{Sohl-Dickstein et~al.(2015)Sohl-Dickstein, Weiss, Maheswaranathan,
  and Ganguli}]{sohl2015deep}
Sohl-Dickstein, J.; Weiss, E.; Maheswaranathan, N.; and Ganguli, S. 2015.
\newblock Deep unsupervised learning using nonequilibrium thermodynamics.
\newblock In \emph{ICML}.

\bibitem[{Sohn, Lee, and Yan(2015)}]{sohn2015learning}
Sohn, K.; Lee, H.; and Yan, X. 2015.
\newblock Learning structured output representation using deep conditional
  generative models.
\newblock In \emph{NeurIPS}.

\bibitem[{Sun and Wu(2019)}]{sun2019image}
Sun, W.; and Wu, T. 2019.
\newblock Image synthesis from reconfigurable layout and style.
\newblock In \emph{ICCV}.

\bibitem[{Sylvain et~al.(2021)Sylvain, Zhang, Bengio, Hjelm, and
  Sharma}]{sylvain2021object}
Sylvain, T.; Zhang, P.; Bengio, Y.; Hjelm, R.~D.; and Sharma, S. 2021.
\newblock Object-centric image generation from layouts.
\newblock In \emph{AAAI}.

\bibitem[{Tan, Shen, and Zhou(2020)}]{tan2020improving}
Tan, S.; Shen, Y.; and Zhou, B. 2020.
\newblock Improving the fairness of deep generative models without retraining.
\newblock \emph{arXiv preprint arXiv:2012.04842}.

\bibitem[{Tang et~al.(2022)Tang, Gu, Bao, Chen, and Wen}]{tang2022improved}
Tang, Z.; Gu, S.; Bao, J.; Chen, D.; and Wen, F. 2022.
\newblock Improved Vector Quantized Diffusion Models.
\newblock \emph{arXiv preprint arXiv:2205.16007}.

\bibitem[{Tao et~al.(2022)Tao, Tang, Wu, Jing, Bao, and Xu}]{tao2022df}
Tao, M.; Tang, H.; Wu, F.; Jing, X.-Y.; Bao, B.-K.; and Xu, C. 2022.
\newblock DF-GAN: A Simple and Effective Baseline for Text-to-Image Synthesis.
\newblock In \emph{CVPR}.

\bibitem[{Vahdat, Kreis, and Kautz(2021)}]{vahdat2021score}
Vahdat, A.; Kreis, K.; and Kautz, J. 2021.
\newblock Score-based generative modeling in latent space.
\newblock In \emph{NeurIPS}.

\bibitem[{Van~den Oord et~al.(2016)Van~den Oord, Kalchbrenner, Espeholt,
  Vinyals, Graves et~al.}]{van2016conditional}
Van~den Oord, A.; Kalchbrenner, N.; Espeholt, L.; Vinyals, O.; Graves, A.;
  et~al. 2016.
\newblock Conditional image generation with pixelcnn decoders.
\newblock In \emph{NeurIPS}.

\bibitem[{Van Den~Oord, Vinyals et~al.(2017)}]{van2017neural}
Van Den~Oord, A.; Vinyals, O.; et~al. 2017.
\newblock Neural discrete representation learning.
\newblock In \emph{NeurIPS}.

\bibitem[{Van~Oord, Kalchbrenner, and Kavukcuoglu(2016)}]{van2016pixel}
Van~Oord, A.; Kalchbrenner, N.; and Kavukcuoglu, K. 2016.
\newblock Pixel recurrent neural networks.
\newblock In \emph{ICML}.

\bibitem[{Wang et~al.(2004)Wang, Bovik, Sheikh, and Simoncelli}]{wang2004image}
Wang, Z.; Bovik, A.~C.; Sheikh, H.~R.; and Simoncelli, E.~P. 2004.
\newblock Image quality assessment: from error visibility to structural
  similarity.
\newblock \emph{TIP}.

\bibitem[{Xu et~al.(2018)Xu, Zhang, Huang, Zhang, Gan, Huang, and
  He}]{xu2018attngan}
Xu, T.; Zhang, P.; Huang, Q.; Zhang, H.; Gan, Z.; Huang, X.; and He, X. 2018.
\newblock Attngan: Fine-grained text to image generation with attentional
  generative adversarial networks.
\newblock In \emph{CVPR}.

\bibitem[{Yan et~al.(2021)Yan, Zhang, Abbeel, and Srinivas}]{yan2021videogpt}
Yan, W.; Zhang, Y.; Abbeel, P.; and Srinivas, A. 2021.
\newblock Videogpt: Video generation using vq-vae and transformers.
\newblock \emph{arXiv preprint arXiv:2104.10157}.

\bibitem[{Yang et~al.(2021)Yang, Fan, Yang, and
  Wang}]{yang2021layouttransformer}
Yang, C.-F.; Fan, W.-C.; Yang, F.-E.; and Wang, Y.-C.~F. 2021.
\newblock Layouttransformer: Scene layout generation with conceptual and
  spatial diversity.
\newblock In \emph{CVPR}.

\bibitem[{Yang et~al.(2022)Yang, Liu, Wang, Yang, and Tao}]{yang2022modeling}
Yang, Z.; Liu, D.; Wang, C.; Yang, J.; and Tao, D. 2022.
\newblock Modeling Image Composition for Complex Scene Generation.
\newblock In \emph{CVPR}.

\bibitem[{Yu et~al.(2015)Yu, Seff, Zhang, Song, Funkhouser, and
  Xiao}]{yu2015lsun}
Yu, F.; Seff, A.; Zhang, Y.; Song, S.; Funkhouser, T.; and Xiao, J. 2015.
\newblock Lsun: Construction of a large-scale image dataset using deep learning
  with humans in the loop.
\newblock \emph{arXiv preprint arXiv:1506.03365}.

\bibitem[{Yu et~al.(2022)Yu, Li, Koh, Zhang, Pang, Qin, Ku, Xu, Baldridge, and
  Wu}]{yu2021vector}
Yu, J.; Li, X.; Koh, J.~Y.; Zhang, H.; Pang, R.; Qin, J.; Ku, A.; Xu, Y.;
  Baldridge, J.; and Wu, Y. 2022.
\newblock Vector-quantized image modeling with improved vqgan.
\newblock In \emph{ICML}.

\bibitem[{Zhang et~al.(2021)Zhang, Koh, Baldridge, Lee, and
  Yang}]{zhang2021cross}
Zhang, H.; Koh, J.~Y.; Baldridge, J.; Lee, H.; and Yang, Y. 2021.
\newblock Cross-modal contrastive learning for text-to-image generation.
\newblock In \emph{CVPR}.

\bibitem[{Zhang et~al.(2017)Zhang, Xu, Li, Zhang, Wang, Huang, and
  Metaxas}]{zhang2017stackgan}
Zhang, H.; Xu, T.; Li, H.; Zhang, S.; Wang, X.; Huang, X.; and Metaxas, D.~N.
  2017.
\newblock Stackgan: Text to photo-realistic image synthesis with stacked
  generative adversarial networks.
\newblock In \emph{ICCV}.

\bibitem[{Zhang et~al.(2018)Zhang, Xu, Li, Zhang, Wang, Huang, and
  Metaxas}]{zhang2018stackgan++}
Zhang, H.; Xu, T.; Li, H.; Zhang, S.; Wang, X.; Huang, X.; and Metaxas, D.~N.
  2018.
\newblock Stackgan++: Realistic image synthesis with stacked generative
  adversarial networks.
\newblock \emph{TPAMI}.

\bibitem[{Zhou et~al.(2022)Zhou, Zhang, Chen, Li, Tensmeyer, Yu, Gu, Xu, and
  Sun}]{zhou2021lafite}
Zhou, Y.; Zhang, R.; Chen, C.; Li, C.; Tensmeyer, C.; Yu, T.; Gu, J.; Xu, J.;
  and Sun, T. 2022.
\newblock LAFITE: Towards Language-Free Training for Text-to-Image Generation.
\newblock In \emph{CVPR}.

\bibitem[{Zhu et~al.(2019)Zhu, Pan, Chen, and Yang}]{zhu2019dm}
Zhu, M.; Pan, P.; Chen, W.; and Yang, Y. 2019.
\newblock Dm-gan: Dynamic memory generative adversarial networks for
  text-to-image synthesis.
\newblock In \emph{CVPR}.

\end{thebibliography}

\clearpage

\renewcommand\thetable{\Alph{table}}
\renewcommand\thefigure{\Alph{figure}}
\renewcommand\thesection{\Alph{section}}
\setcounter{figure}{0}   
\setcounter{table}{0}   
\setcounter{section}{0}   

\begin{figure*}[!ht]
  \centering
  \includegraphics[page=5,trim={910 880 920 880}, clip, width=0.90\textwidth]{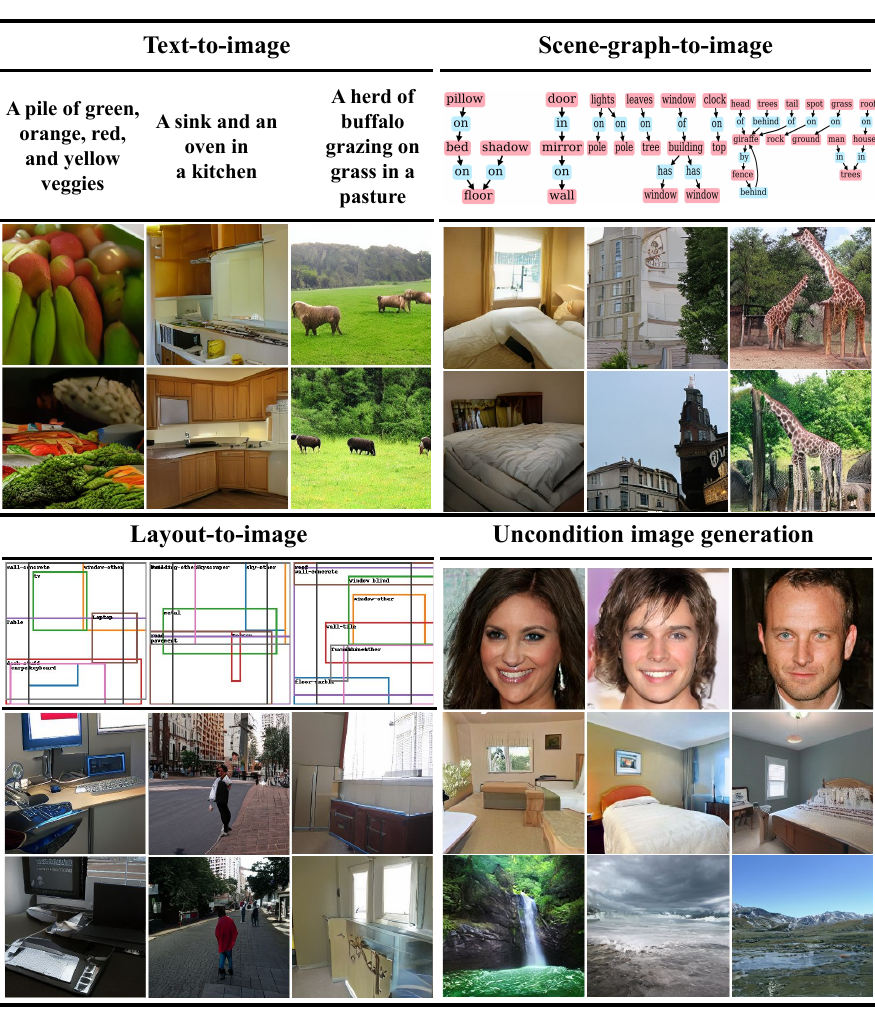}
  \vspace{-3mm}
  \caption{Generated examples of \our on various tasks. Full-scale version of Fig.~\textcolor{black}{2} in the main paper.}
  \vspace{-4mm}
  \label{demo:big_full}
\end{figure*}

\section{Diffusion and Denoising Details}

\subsection{Formulation of the Diffusion and Denoising Process of \our}

With the deployment of the MS-VQGAN mentioned in Section~\textcolor{blue}{3.1}, \our is able to encode the input image into feature maps $\mathcal{Z}$ with different spatial levels. We then introduce feature pyramid diffusion model (\our) to further learn the distribution of the latent space and generate images from sampled normal distribution variables. Similar to other diffusion methods (DMs), our \our contains two parts: the \textit{diffusion process} and the \textit{denoising process}. We now detail these two parts in the following paragraphs.

\subsubsection{Diffusion Process of \our}
The diffusion process aims at producing the approximate posterior $q(\vectorbold{Z}_{1:T}|\vectorbold{z}_{0})$, which is fixed to a Markov chain that gradually adds Gaussian noise to the input data based on a fixed schedule $\beta_1, ..., \beta_T$. As shown in Figure~\textcolor{blue}{3}, instead of directly adding noise on the multi-scale feature maps $\mathcal{Z} = \{\vectorbold{z}^1, \vectorbold{z}^2, ..., \vectorbold{z}^N\}$, we conduct diffusion process on each scale of feature map sequentially from low-level one (i.e., $\vectorbold{z}^1$) to high-level one (i.e., $\vectorbold{z}^N$) over the course of N $\times$ T timesteps. That is, the approximate posterior q can be formulated as follows. 

\begin{equation}
    \begin{aligned}
        q(\vectorbold{z}^{1:N}_{1:T}|\mathcal{Z})&=\prod\limits_{n=1}^{N}\prod\limits_{t=1}^{T}q(\vectorbold{z}^n_t|\vectorbold{z}^n_{t-1}),\text{ and } \\ & q(\vectorbold{z}^n_t|\vectorbold{z}^n_{t-1})=\mathcal{N}(\sqrt{1-\beta_{t}}\vectorbold{z}^n_{t-1}, \beta_t \vectorbold{I}),
    \end{aligned}
\end{equation}
where noise schedules $\beta$ can be fixed or learnable parameters. Different from the classical diffusion process that corrupts pixels into noise in an unbiased way, we observe that \our’s diffusion process starts from corrupting the object details, object shape, and finally the structure of the entire image. This allows Frido to capture information in different semantic levels.

\subsubsection{Denoising Process of \our}
The denoising process starts from the joint distribution $p(\vectorbold{z}^{1:N}_T) \sim \mathcal{N}(\vectorbold{z}^{1:N}_T;\textbf{0}, \textbf{I})$. That is, feature maps of all scales are initialized from a normal distribution. Different from classical denoising process as Eq.~\textcolor{blue}{2}, we conduct the denoising process from high-level feature maps $p(\vectorbold{z}^N_T)$ to low-level feature map $p(\vectorbold{z}^1_T)$ sequentially, achieving coarse-to-fine generation. In other words, the denoising process for $n$-th scale would be based on the previous produced feature maps $\vectorbold{z}^N_0, \vectorbold{z}^{N-1}_0, ..., \vectorbold{z}^{n+1}_0$ and can be formulated as $p_\theta(\vectorbold{z}^n_0|\vectorbold{z}^{n+1:N}_0) = \int p_\theta(\vectorbold{z}^n_{0:T}|\vectorbold{z}^{n+1:N})d\vectorbold{z}^n_{1:T}$, where

\begin{equation}
    \begin{aligned}
        p_\theta&(\vectorbold{z}^n_{0:T}|\vectorbold{z}^{n+1:N})
        = p(\vectorbold{z}^n_{T})\prod\limits_{t=1}^{T}p_\theta(\vectorbold{z}^n_{t-1}|\vectorbold{z}^n_{t}, \vectorbold{z}^{n+1:N}),\text{ and } \\ 
        & p_\theta(\vectorbold{z}_{t-1}|\vectorbold{z}_{t}, \vectorbold{z}^{n+1:N})=\\
        &\mathcal{N}(\vectorbold{z}_{t-1};\mu_\theta(\vectorbold{z}_t, t, \vectorbold{z}^{n+1:N}), \sigma_\theta(\vectorbold{z_t}, t, \vectorbold{z}^{n+1:N})).
    \end{aligned}
\end{equation}

Practically, a designed denoising autoencoder $\epsilon_\theta$ is adopted to we can simplify our model as a sequence of denoising autoencoder $\epsilon_\theta(\vectorbold{z}^n_t, \vectorbold{z}^{n+1:N}_0, t)$, $t=1...T$ and $n = N...1$, with the goal of predicting the added noise variant on the $\vectorbold{z}^n_t$. We then adopt our feature pyramid U-Net (PyU-Net) as the denoising autoencoder $\epsilon_\theta$ and the objective function can be formulated as follows. (modified from Eq.\textcolor{blue}{3})

\begin{equation}
    \begin{aligned}
        \mathcal{L}_{\text{\our}} =& \\ \mathbb{E}_{\vectorbold{z}^n_0, \epsilon, t} &\left[ \|\epsilon-\epsilon_\theta(\sqrt{\Tilde{\alpha_t}}\vectorbold{z}^n_0+
        \sqrt{1-\Tilde{\alpha_t}}\epsilon, \vectorbold{z}^{n+1:N}_0)\|^2  \right],
    \end{aligned}
\end{equation}
where t is uniformly sampled from {1, . . . , T}, and n is uniformly sampled from {1, . . . , N}. In the testing phase, the samples from $p(\vectorbold{z}^{1:N}_0)$ can be decoded back to image pixel space with the pre-trained decoder $\mathcal{D}$.

\section{Implementation Details}
For the reproducibility of \our, we detail the implementation details for each task, including but not limit to the training/testing environment, hyper-parameters setting for each task, and the details of encoding multi-modal conditions.

\begin{table*}[]
\centering
\resizebox{0.95\textwidth}{!}{
\begin{tabular}{cccccc}
\multicolumn{1}{c|}{Hyper-parameter} & T2I (Frido-f16f8) & SG2I (Frido-f16f8) & Label2I (Frido-f16f8) & Layout2I (Frido-f16f8) & Layout-to-image (Frido-f8f4) \\ \hline
\multicolumn{6}{c}{General settings} \\ \hline
\multicolumn{1}{c|}{Base learning rate} & 2e-7 & 2e-7 & 1e-6 & 2e-7 & 1e-7 \\
\multicolumn{1}{c|}{Scale lr} & True & True & False & True & True \\
\multicolumn{1}{c|}{Batch size} & 32 & 32 & 32 & 32 & 12 \\
\multicolumn{1}{c|}{Training epochs} & 300 & 250/150 (COCO/VG) & 150 & 500 & 200/25 (COCO/IO) \\
\multicolumn{1}{c|}{MS-VQGAN} & MS-VQGAN-f16f8 & MS-VQGAN-f16f8 & MS-VQGAN-f16f8 & MS-VQGAN-f16f8 & MS-VQGAN-f8f4 \\
\multicolumn{1}{c|}{Diffusion steps} & 1000 & 1000 & 1000 & 1000 & 1000 \\
\multicolumn{1}{c|}{Noise Schedule} & linear & linear & linear & linear & linear \\
\multicolumn{1}{l|}{Number of feature scales} & 2 & 2 & 2 & 2 & 2 \\ \hline
\multicolumn{6}{c}{PyU-Net settings} \\ \hline
\multicolumn{1}{c|}{Model channels} & 192 & 192 & 192 & 192 & 192 \\
\multicolumn{1}{c|}{Number of res blocks} & 2 & 2 & 2 & 2 & 2 \\
\multicolumn{1}{c|}{Attention resolutions} & 2, 4, 8 & 2, 4, 8 & 2, 4, 8 & 2, 4, 8 & 2, 4, 8 \\
\multicolumn{1}{c|}{Channel multiplier} & 1, 2, 3, 5 & 1, 2, 3, 5 & 1, 2, 3, 5 & 1, 2, 3, 5 & 1, 2, 3, 5 \\
\multicolumn{1}{c|}{Transformer depth} & 1 & 1 & 1 & 1 & 1 \\
\multicolumn{1}{c|}{Number of heads} & 6 & 6 & 6 & 6 & 6 \\
\multicolumn{1}{c|}{Context dim} & 640 & 640 & 640 & 640 & 640 \\
\multicolumn{1}{c|}{Noise ratio} & 0.1 & 0.1 & 0.1 & 0.1 & 0.1 \\ \hline
\multicolumn{6}{c}{Multi-modal Condition Encoding settings} \\ \hline
\multicolumn{1}{c|}{Encoding module} & BERTEmbedder & BERTEmbedder & BERTEmbedder & BERTEmbedder & BERTEmbedder \\
\multicolumn{1}{c|}{Number of layer} & 32 & 32 & 32 & 32 & 32 \\
\multicolumn{1}{c|}{Attention embedding dim} & 640 & 640 & 640 & 640 & 640 \\
\multicolumn{1}{c|}{Vocab size} & 30522 & 30522 & 30522 & 30522 & 30522 \\
\multicolumn{1}{c|}{Max sequence length} & 77 & 180 & 77 & 96 & 96
\end{tabular}}
\caption{Hyper-parameter settings for all \our of all conditional tasks, including text-to-image (T2I), scene-graph-to-image (SG2I), label-to-image (Label2I), and layout-to-image (Layout2I) in this paper. IO denotes Open-Image dataset. If model converges earlier, we used the checkpoint with best total ema loss. Note that, for \our-f8f4 on COCO, model was fine-tuned from the one trained on OI.}
\label{hyper:cond}
\end{table*}

\begin{table*}[t]
\centering
\resizebox{0.65\textwidth}{!}{
\begin{tabular}{cccc}
\multicolumn{1}{c|}{Hyper-parameter} & CelebA-HQ (Frido-f8f4) & Landscape (Frido-f8f4) & LSUN-bed (Frido-f8f4) \\ \hline
\multicolumn{4}{c}{General settings} \\ \hline
\multicolumn{1}{c|}{Base learning rate} & 1e-6 & 3e-7 & 3e-7 \\
\multicolumn{1}{c|}{Scale lr} & False & True & True \\
\multicolumn{1}{c|}{Batch size} & 8 & 6 & 6 \\
\multicolumn{1}{c|}{Training epochs} & 350 & 300 & 50 \\
\multicolumn{1}{c|}{MS-VQGAN} & MS-VQGAN-f8f4 & MS-VQGAN-f8f4 & MS-VQGAN-f8f4 \\
\multicolumn{1}{c|}{Diffusion steps} & 1000 & 1000 & 1000 \\
\multicolumn{1}{c|}{Noise scheduler} & linear & linear & linear \\
\multicolumn{1}{l|}{Number of feature scales} & 2 & 2 & 2 \\ \hline
\multicolumn{4}{c}{PyU-Net settings} \\ \hline
\multicolumn{1}{c|}{Model channels} & 224 & 192 & 192 \\
\multicolumn{1}{c|}{Number of res blocks} & 2 & 2 & 2 \\
\multicolumn{1}{c|}{Attention resolutions} & 2, 4, 8 & 1, 2, 4 & 1, 2, 4 \\
\multicolumn{1}{c|}{Channel multiplier} & 1, 2, 3, 4 & 1, 2, 4 & 1, 2, 4 \\
\multicolumn{1}{c|}{Number of heads} & 7 & 1 & 1 \\
\multicolumn{1}{c|}{Noise ratio} & 0.1 & 0.1 & 0.1
\end{tabular}}
\caption{Hyper-parameter settings for all \our of all unconditional tasks, including unconditional image generation on CelebA-HQ, Landscape, and LSUN-bed. If model converges earlier, we used the checkpoint with best total ema loss.}
\label{hyper:uncond}
\end{table*}

\begin{table*}[]
\centering
\resizebox{0.94\textwidth}{!}{
\begin{tabular}{ccccccc}
\multicolumn{1}{c|}{Hyper-parameter} & \begin{tabular}[c]{@{}c@{}}Model ablation \\ (Fig.6)\end{tabular} & \begin{tabular}[c]{@{}c@{}}Noise analysis \\ (Fig.B)\end{tabular} & \begin{tabular}[c]{@{}c@{}}Multi-scale analysis \\ (Frido-f16f8, Table~\ref{exp:scale_analysis})\end{tabular} & \begin{tabular}[c]{@{}c@{}}Multi-scale analysis \\ (Frido-f8f4, Table~\ref{exp:scale_analysis})\end{tabular} & \begin{tabular}[c]{@{}c@{}}Multi-scale analysis \\ (Frido-f32f16f8, Table~\ref{exp:scale_analysis})\end{tabular} & \begin{tabular}[c]{@{}c@{}}Multi-scale analysis \\ (Frido-f16f8f4, Table~\ref{exp:scale_analysis})\end{tabular} \\ \hline
\multicolumn{7}{c}{General settings} \\ \hline
\multicolumn{1}{c|}{Base learning rate} & 2e-7 & 4e-6 & 2e-7 & 2e-7 & 2e-7 & 2e-7 \\
\multicolumn{1}{c|}{Scale lr} & True & False & True & True & True & True \\
\multicolumn{1}{c|}{Batch size} & 32 & 32 & 32 & 6 & 12 & 2 \\
\multicolumn{1}{c|}{Training epochs} & 120 & 120 (COCO/VG) & 120 & 120 & 120 & 120 \\
\multicolumn{1}{c|}{MS-VQGAN} & \begin{tabular}[c]{@{}c@{}}MS-VQGAN\\ f16f8\end{tabular} & \begin{tabular}[c]{@{}c@{}}MS-VQGAN\\ f16f8\end{tabular} & \begin{tabular}[c]{@{}c@{}}MS-VQGAN\\ f16f8\end{tabular} & \begin{tabular}[c]{@{}c@{}}MS-VQGAN\\ f8f4\end{tabular} & \begin{tabular}[c]{@{}c@{}}MS-VQGAN\\ f32f16f8\end{tabular} & \begin{tabular}[c]{@{}c@{}}MS-VQGAN\\ f16f8f4\end{tabular} \\
\multicolumn{1}{c|}{Diffusion steps} & 1000 & 1000 & 1000 & 1000 & 1000 & 1000 \\
\multicolumn{1}{c|}{Noise Schedule} & linear & linear & linear & linear & linear & linear \\
\multicolumn{1}{l|}{Number of feature scales} & 2 & 2 & 2 & 2 & 3 & 3 \\ \hline
\multicolumn{7}{c}{PyU-Net settings} \\ \hline
\multicolumn{1}{c|}{Model channels} & 192 & 192 & 192 & 192 & 192 & 192 \\
\multicolumn{1}{c|}{Number of res blocks} & 2 & 2 & 2 & 2 & 2 & 2 \\
\multicolumn{1}{c|}{Attention resolutions} & 2, 4, 8 & 2, 4, 8 & 2, 4, 8 & 2, 4, 8 & 2, 4, 8 & 2, 4, 8 \\
\multicolumn{1}{c|}{Channel multiplier} & 1, 2, 3, 5 & 1, 2, 3, 5 & 1, 2, 3, 5 & 1, 2, 3, 5 & 1, 2, 3, 5 & 1, 2, 3, 5 \\
\multicolumn{1}{c|}{Transformer depth} & 1 & 1 & 1 & 1 & 1 & 1 \\
\multicolumn{1}{c|}{Number of heads} & 6 & 6 & 6 & 6 & 6 & 6 \\
\multicolumn{1}{c|}{Context dim} & 640 & 640 & 640 & 640 & 640 & 640 \\
\multicolumn{1}{c|}{Noise ratio} & 0.1 & 0.1 & 0.1 & 0.1 & 0.1 & 0.1 \\ \hline
\multicolumn{7}{c}{Multi-modal Condition Encoding settings} \\ \hline
\multicolumn{1}{c|}{Encoding module} & BERTEmbedder & BERTEmbedder & BERTEmbedder & BERTEmbedder & BERTEmbedder & BERTEmbedder \\
\multicolumn{1}{c|}{Number of layer} & 32 & 32 & 32 & 32 & 32 & 32 \\
\multicolumn{1}{c|}{Attention embedding dim} & 640 & 640 & 640 & 640 & 640 & 640 \\
\multicolumn{1}{c|}{Vocab size} & 30522 & 30522 & 30522 & 30522 & 30522 & 30522 \\
\multicolumn{1}{c|}{Max sequence length} & 77 & 77 & 77 & 77 & 77 & 77
\end{tabular}}
\caption{Hyper-parameter settings for \our of all model analysis tasks. If model converges earlier, we used the checkpoint with best total ema loss. }
\label{hyper:ablation}
\end{table*}

\subsection{Training/Testing Environment}
In this paper, if not specified, models are trained on a Linux environment with 8 Nvidia Tesla V100. Also, the training and testing are powered by Pytorch Lightning~\cite{pytorchLightning} of 1.4.2, with operational precision as float32. While training, models are run with raw batch size on each GPU without the gradient accumulation trick. For better reproducibility, we use the ``\textit{seed}$\_$\textit{everything}" tools in Pytorch Lightning to fix the random seeds in the training phase with seed$=$23. For the optimization setting, we use the AdamW algorithm as the optimizer with the default setting as follows: betas$=(0.9, 0.999)$, eps$=1e-08$, weight decay$=0.01$, and learning rates are specified in the later hyper-parameters setting section. A learning rate scheduler is not used in the training of our methods, if not specified.

\subsection{Hyper-Parameter Settings}
For a better understanding of the hyper-parameter setting of \our for each task, we provide an overview of hyper-parameter settings of all \our models in this paper. For conditional image generation tasks, the hyper-parameter settings are listed in Table~\ref{hyper:cond}. For unconditional image generation, we detail the settings in Table~\ref{hyper:uncond}. As for modal analysis, the settings of the hyper-parameter are provided in Table~\ref{hyper:ablation}.

\subsection{Encoding Multi-Modal Conditions}

\subsubsection{Text-to-image Generation}
Similar to previous diffusion works~\cite{rombach2022high, nichol2021glide}, we use BERT Embedder to convert a sequence of text conditions into a sequence of embeddings. Specifically, given a caption as a cross-modal condition, a BERT tokenizer is deployed to tokenize a caption into a sequence of tokens, followed by a BERT transformer module to encode these tokens to a sequence of embeddings.

\subsubsection{Image Generation from Scene Graph}
Similar to LT-Net~\cite{yang2021layouttransformer}, we first transform the scene graph conditions into three sequences of tokens, including the \textit{word tokens}, \textit{part-of-pair tokens}~\cite{yang2021layouttransformer}, and \textit{object ID} tokens.
After that, the final condition embedding sequence can be obtained by first adopting BERT-tokenizer~\cite{devlin2018bert} to encode these tokens into embeddings and sum up these embeddings.

\subsubsection{Layout-to-Image Generation}
To encode a sequence of image-level labels into an embedding sequence, we first map the labels to the corresponding category names. After that, these class names are concatenated as a text sequence and encoded by a similar encoding approach used in text-to-image generation.

\subsubsection{Layout-to-Image Generation}
We follow the layout discretization process in LDM~\cite{rombach2022high} to transform the layout of an image into a sequence of tokens, where conditions are discretized into multiple triples. Each triple represents the spatial locations, size, and category of the corresponding bounding box. More specifically, each bounding box is encoded as a (l, b, c)-tuple, where l denotes the top-left and b the bottom-right position. c represents the category of the bounding box. With a sequence of tokens obtained, we adopt a BERT transformer encoder to encode the sequence.

\section{Datasets and Evaluations}
\subsection{Evaluation Metrics}
For quantitative experiments, we consider various metrics from different aspects to evaluate our method on each task. We now introduce these metrics as follows:

\begin{figure}[t]
  \centering
  \includegraphics[page=15,trim={180 35 150 40}, clip, width=0.45\textwidth]{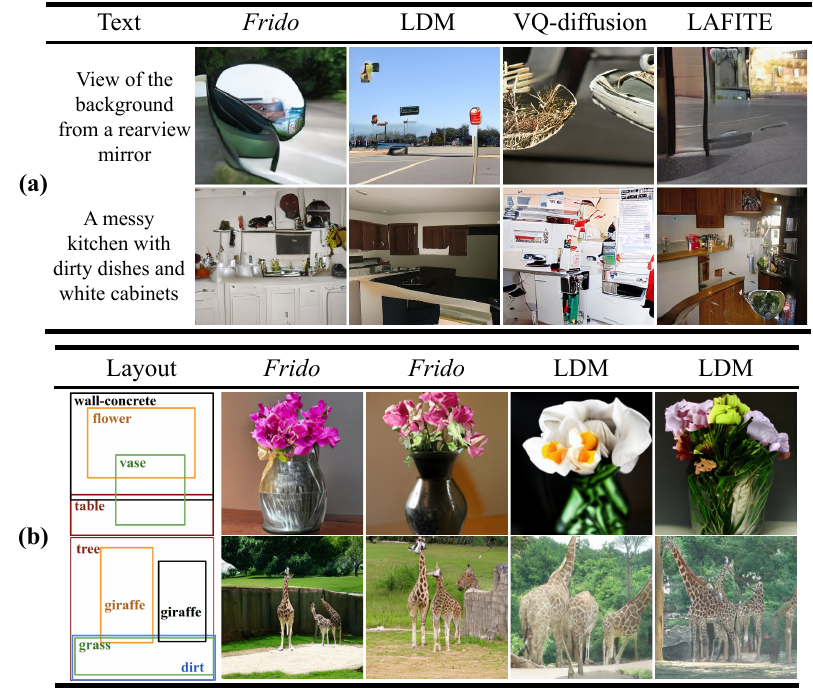}
  \vspace{-1mm}
  \caption{(a) Qualitative comparison on T2I. (b) Qualitative comparison on the Layout2I generation. Note that the layouts are obtained from the LDM paper~\cite{rombach2022high}. Please see the supplementary for more tasks and examples.}
  \vspace{-2mm}
  \label{demo:T2I}
\end{figure}
\begin{figure}[t]
  \centering
  \includegraphics[page=24,trim={160 0 175 5}, clip, width=0.45\textwidth]{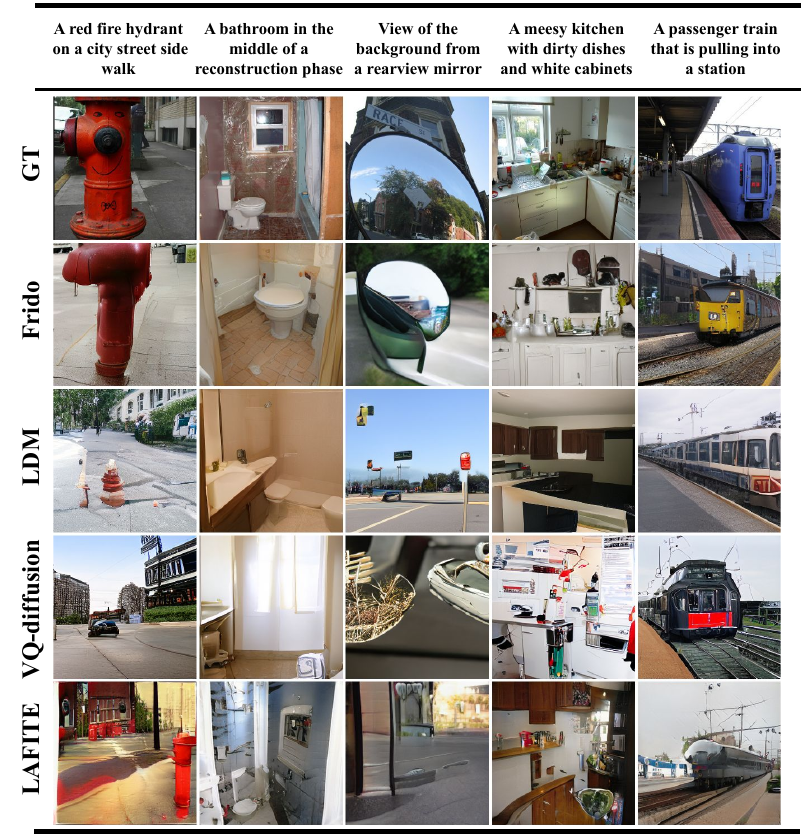}
  \vspace{-0mm}
  \caption{More qualitative comparison on T2I on COCO. Note that the first row denotes the text conditions.}
  \vspace{-2mm}
  \label{demo:T2I}
\end{figure}
\begin{figure}[t]
  \centering
  \includegraphics[page=22,trim={165 10 165 10}, clip, width=0.45\textwidth]{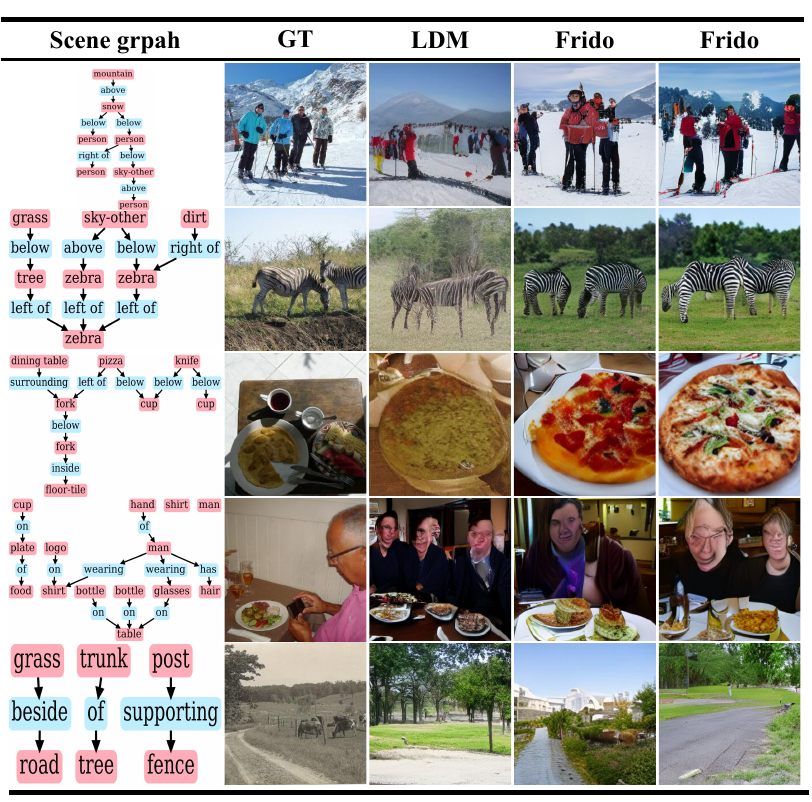}
  \vspace{0mm}
  \caption{More qualitative comparison on scene-graph-to-image generation.}
  \vspace{-2mm}
  \label{demo:SG2I}
\end{figure}

\begin{figure}[t]
  \centering
  \includegraphics[page=25,trim={150 40 150 40}, clip, width=0.45\textwidth]{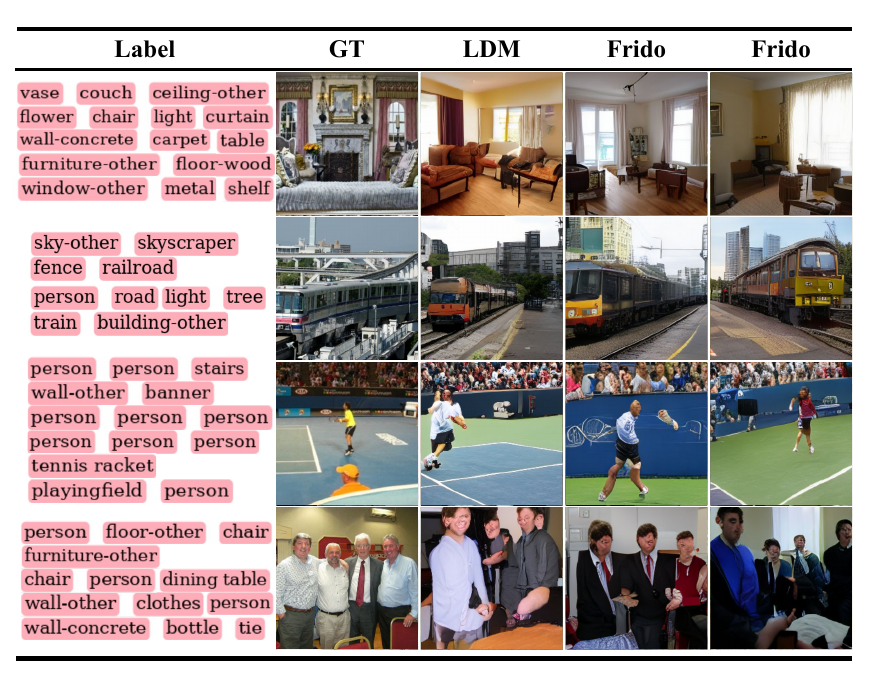}
  \vspace{0mm}
  \caption{More qualitative comparison on the label-to-image generation.}
  \vspace{-2mm}
  \label{demo:Label2I}
\end{figure}

\begin{figure}[t]
  \centering
  \includegraphics[page=23,trim={185 45 185 45}, clip, width=0.45\textwidth]{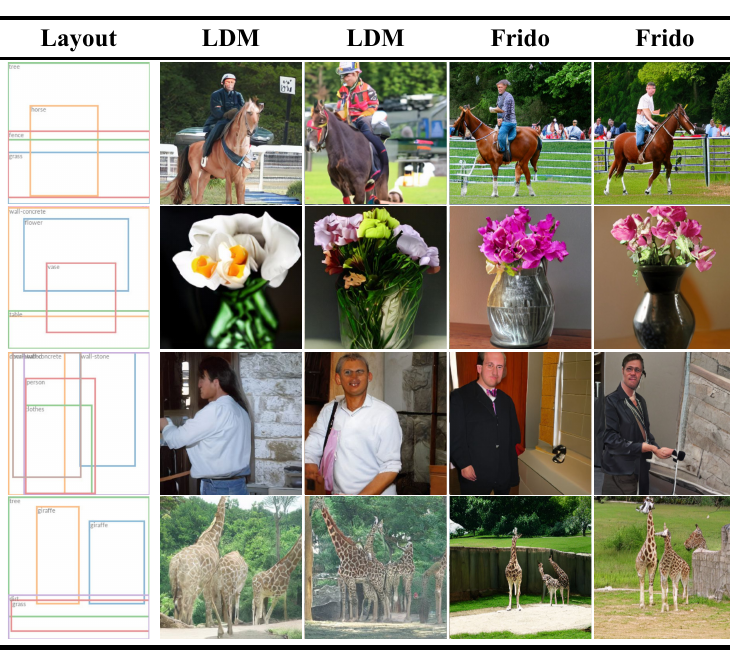}
  \vspace{0mm}
  \caption{More qualitative comparison on the layout-to-image generation. Note that the layouts are from the paper of LDM~\cite{rombach2022high}.}
  \vspace{-2mm}
  \label{demo:Layout2I}
\end{figure}

\begin{itemize}
  \item To measure the image quality, we consider Fréchet inception distance (FID) and Inception score (IS). IS evaluates the distribution of generated images by an Inception-V3 model pre-trained on ImageNet. Unlike the inception score, FID takes the real image used to train the generative model into consideration and compares the distribution of that with the distribution of generated images, making it a widely-used metric to access the image quality. Practically, we use \textit{torch-fidelity}\footnote{torch-fidelity: https://github.com/toshas/torch-fidelity} to calculate the FID and inception score. Furthermore, sceneFID~\cite{sylvain2021object} is considered to measure the image quality at instance-level. To calculate SceneFID, we follow ~\citet{sylvain2021object} to first crop the generated images by the corresponding ground truth bounding boxes, yielding the generated instances. After that, similar to FID score, we use \textit{torch-fidelity} to calculate the FID score between ground truth instances and the generated instances. 
  
  \item In order to perform the quantitative analysis on the semantic correctness of text-to-image or scene-graph-to-image generation, we take into account of reference-free CLIP score (CLIPscore)~\cite{hessel2021clipscore} as the measurement for these tasks. CLIPsocre adopt CLIP~\cite{radford2021learning} pre-trained on ~400M image-caption pairs from the web dataset to assess image-caption compatibility. In this paper, we use the officially released code\footnote{clipscore: https://github.com/jmhessel/clipscore} to calculate the CLIP score.
  
  \item YOLO detection score~\cite{li2021image} is designed to measure the alignment and fidelity of the generated objects in Layout-to-image task. We simply follow the setting in LAMA and TwFA~\cite{yang2022modeling} and adopt YOLOv4~\cite{bochkovskiy2020yolov4} as the detector. In detail, we use the officially released COCO pre-trained YOLOv4 model weights in the evaluation and report AP (average precision) with 0.50 threshold in this paper.
  
  \item Precision and Recall for image generation task~\cite{sajjadi2018assessing, kynkaanniemi2019improved} is to measure the distribution coverage of a trained generative model. In other words, precision is the probability that a sample from a trained generative model lands in the support of the distribution of the dataset, and recall is the probability that a sample from the dataset is within the support of the distribution of the trained generative model. Therefore, a generative model with high precision is able to generate a sample with high realism as the real data, and a high recall denotes the distribution learned by the generative model covers most part of the real distribution, resulting in high diversity generated results. Note that we adopt the evaluation code\footnote{guided-diffusion: https://github.com/openai/guided-diffusion} of guided-diffusion~\cite{dhariwal2021diffusion} to compute the precision and recall for generative models.
  \item For the analysis of the multi-scale vector quantization process, we consider the classical Peak signal-to-noise ratio (PSNR) and structural index similarity (SSIM)~\cite{wang2004image} as the reconstruction quality assessment, which are common metrics for evaluating the similarity between images. Also, we use \textit{pytorch-gan-metrics}\footnote{gan-metrics: https://pypi.org/project/pytorch-gan-metrics/} to compute these metrics.
  
\end{itemize}

\subsection{Datasets and Evaluation Setting}
In this section, for reproducibility, we detail the evaluation setting for each task on each dataset.

\begin{itemize}
    \item Text-to-image generation (T2I): We perform experiments of T2I on standard COCO setting~\cite{zhou2021lafite}. In the standard T2I setting, models are trained on COCO 2014 captioning dataset only and evaluated on COCO 2014 validation set. In COCO 2014, the numbers of training and validation data are 83K and 41k, respectively. We note that, in the evaluation phase, models are evaluated on the full COCO 2014 validation set.
    
    \item Scene-graph-to-image generation (SG2I): We conduct experiments of SG2I on COCO and Visual Genome datasets. For the COCO track, following the setting of sg2im~\cite{johnson2018image} and Canonicalsg2im~\cite{herzig2020learning}, we train our model on COCO 2017 Stuff Segmentation Challenge split, containing 25K/2K training/testing images. Also, we adopt the same pre-processing in sg2im to obtain the scene graph for each training image. In the Visual Genome dataset, the scene-graph-image pairs are given. In this paper, we use the same subset in sg2im, which ignores tiny objects and only uses images with 3~30 objects and 1+ relationships. This results in  62K/5K training/testing data. For the quantitative comparison, we also follow the evaluation protocol of previous works~\cite{herzig2020learning, jahn2021high}. 
    
    \item Label-to-image generation: 
    We evaluate the label-to-image models on COCO datasets. Following the evaluation protocol of~\cite{jyothi2019layoutvae, yang2021layouttransformer}, we use the 2017 Panoptic version of COCO, containing 118K/5K training/validation images with bounding box annotations for each image (we only use image-level labels for training). For a better understanding of the performance on the scene image generation of different difficulties, we consider two settings on the validation sets. The first one is images with only 3-8 objects (total of 3,276 images), and the second one is validation images with 2-30 objects (total of 4,722).
    
    \item Layout-to-image generation: Similar to scene-graph-to-image generation, we perform the model training on COCO 2017 Stuff Segmentation Challenge split for a fair comparison with previous works. In the testing phase, we follow the common practice~\cite{rombach2022high, sylvain2021object, jahn2021high} and report the FID and IS on the testing set with 2,048 images. As for the OpenImage dataset, we follow the setting of VQGAN+T~\cite{jahn2021high} and sample 2,048 images in the validation set for evaluation.

    \item Unconditional image generation: We conduct unconditional image generation on LSUN-bed, CelebA-HQ, and Landscape datasets. For all datasets, we train our model with all the training sets without any filtering. In the evaluation phase, we follow common practice~\cite{rombach2022high, anokhin2021image, casanova2021instance} to evaluate the performance on LSUN-bedand, CelebA-HQ, Landscape, where metrics are calculated between the 50k random generated images and the entired training set. Also, we use \textit{torch-fidelity} to compute the FID score and utilize the script provided by ADM~\cite{dhariwal2021diffusion} to obtain precision/recall scores. 
\end{itemize}

\section{Additional Experiment Results}
\subsection{Conditional Image Generation}

In this section, we provide more qualitative comparison results on various conditional image generation tasks, including text-to-image (Figure~\ref{demo:T2I}), scene-graph-to-image (Figure~\ref{demo:SG2I}), label-to-image (Figure~\ref{demo:Label2I}), and layout-to-image (Figure~\ref{demo:Layout2I}). Note that the settings of these qualitative experiments are the same as described in Section~\textcolor{blue}{4.2}.

\begin{table}[t]
\center
\resizebox{0.48\textwidth}{!}{
\setlength{\tabcolsep}{0.5mm}{
\begin{tabular}{l|ccc|cc}
\multirow{2}{*}{Methods} & \multicolumn{3}{c|}{COCO 256} & \multicolumn{2}{c}{Visual Genome 256} \\ \cline{2-6} 
 & FID$\downarrow$ & YOLO$\uparrow$ & SceneFID$\downarrow$ & FID$\downarrow$ & SceneFID$\downarrow$ \\ \hline
LostGAN-V2 & 42.55 & 15.3 & 22.00 & 47.62 & 18.27 \\
OC-GAN & 41.65 & - & - & 40.85 & - \\
HCSS & 33.68 & - & 13.36 & 19.14 & 8.61 \\
LAMA & 31.12 & 19.7 & 18.64 & 31.63 & 13.66 \\
Context-L2I & 29.56 & - & 14.40 & - & - \\
TwFA & 22.15 & - & 11.99 & 17.74 & 7.54 \\
\hline
\our-f8f4 & 25.86 & 28.1 & 11.28 & 20.16 & 8.63 \\
\our-f8f4-G & \textbf{21.67} & \textbf{30.4} & \textbf{10.44} & \textbf{17.24} & \textbf{6.52}
\end{tabular}}}
\vspace{-2mm}
\caption{Layout-to-image generation on COCO-stuff 2017 and Visual Genome.}
\vspace{-3mm}
\label{exp:layout2i_extra}
\end{table}

\subsubsection{More Experimental Results for Layout-to-Image Generation}
Continuing from the Section~\ref{sec:cond}, we follow TwFA~\cite{yang2022modeling} and conduct experiments on standard COCO stuff and Visual Genome datasets. As the results shown in Table~\ref{exp:layout2i_extra}, we compare with the previous approaches, including HCSS~\cite{jahn2021high}, LAMA~\cite{li2021image}, Context-L2I~\cite{he2021context}, and TwFA.
In this table, \our again surpasses previous approaches on both FID and Scene FID, verifying the preferable of \our on the generation of multiple objects' shapes and details.

\begin{table}[t]
\centering
\setlength{\tabcolsep}{3mm}{
\begin{tabular}{l|ccc}
Name & FID & Precision & Recall \\
\hline
DC-VAE & 15.8 & - & - \\
VQGAN+T. (k=400) & 10.2 & - & - \\
PGGAN & 8.0 & - & - \\
LSGM & 7.22 & - & - \\
UDM & 7.16 & - & - \\
LDM-4 & \textbf{5.11} & \textbf{0.72} & \ul{0.49} \\
\hline
Ours-f8f4 & \ul{6.38} & \ul{0.71} & \textbf{0.50}
\end{tabular}}
\caption{Unconditional image generation on CelebA-HQ.}
\label{exp:CelebA-HQ}
\end{table}

\begin{table}[t]
\centering
\setlength{\tabcolsep}{4mm}{
\begin{tabular}{l|ccc}
Name & FID & Precision & Recall \\
\hline
CIPS & \ul{3.61} & - & - \\
StyleGAN-v2 & \textbf{2.36} & - & - \\
LDM-4$^\dagger$ & 7.69 & \textbf{0.69} & \ul{0.49} \\
\hline
PyDM-f8f4 & 5.29 & \ul{0.67} & \textbf{0.51}
\end{tabular}}
\caption{Unconditional image generation on Landscape 256. $^\dagger$ denotes models are trained by official code and configs.}
\label{exp:landscape}
\end{table}

\begin{table}[t]
\centering
\setlength{\tabcolsep}{3mm}{
\begin{tabular}{l|ccc}
Name & FID & Precision & Recall \\
\hline
ImageBART & 5.51 & - & - \\
DDPM & 4.9 & - & - \\
UDM & 4.57 & - & {\color[HTML]{D9D9D9} -} \\
StyleGAN & 2.35 & 0.59 & 0.48 \\
ADM & \ul{1.90} & \textbf{0.66} & \textbf{0.51} \\
ProjectedGAN & \textbf{1.52} & 0.61 & 0.34 \\
LDM-4 & 2.95 & \textbf{0.66} & \ul{0.48} \\
\hline
PyDM-f8f4 & 3.87 & \ul{0.64} & \textbf{0.51}
\end{tabular}}
\caption{Unconditional image generation on LSUN-Beds.}
\label{exp:bed}
\end{table}

\begin{figure}[t]
  \centering
  \includegraphics[page=16,trim={195 45 200 50}, clip, width=0.45\textwidth]{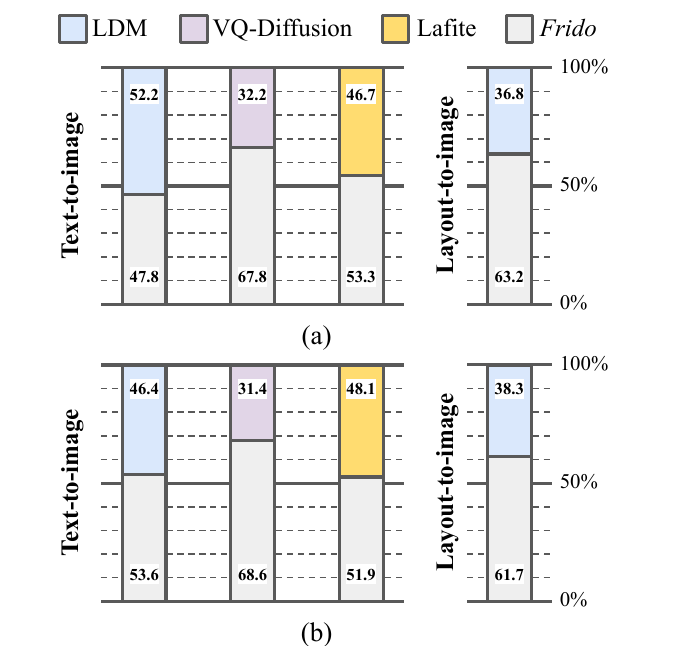}
  \vspace{-1mm}
  \caption{User studies of T2I and Layout2I on COCO. Given images produced by two different models, users are asked to choose the one which (a) exhibits better image fidelity and (b) is generated with higher condition relevance.}
  \vspace{-2mm}
  \label{demo:user}
\end{figure}

\subsection{User Preference Studies}
\label{sec:user_study}

To further provide subjective evaluation of our model, we conduct user studies for T2I and Layout2I with two metrics. Given two generated images produced by two different models, users are asked to evaluate the generated images from two aspects: (1)``which synthesized image is with higher image fidelity regardless of the semantic correctness?" (2) ``which of the generated image has higher condition relevance?" For each comparison, we randomly picked 15 images from the validation set and compared the images generated by our method to those generated by the SOTA of each task, including LDM~\cite{rombach2022high}, VQ-Diffusion~\cite{gu2022vector}, and Lafite~\cite{zhou2021lafite}. In total, we collected 1,320 answers from 22 users

As the results shown in Fig.~\ref{demo:user} (b), we can find that \our outperforms previous approaches by clear margins on the metric of condition relevance. For image fidelity in Fig.~\ref{demo:user} (a), \our surpasses VQ-diffusion and Lafite on T2I and LDM on Layout2I. While LDM produced a higher visual preference on T2I task, we find that it is mainly due to the fact that LDM tends to generate image with high visual smoothness yet low accordance with the constraints. Overall, the above user studies show similar tendency as the quantitative studies in Tables~\ref{exp:T2I} and ~\ref{exp:layout2i}, verifying the preference of \our for text-to-image and layout-to-image generation.

\subsection{Unconditional Image Generation}
To verify the generalizability of \our, We also analyze the performance of our method on the unconditional image generation task, and compare \our with previous works, including DC-VAE~\cite{parmar2021dual}, VQGAN~\cite{esser2021taming}, PGGAN~\cite{karras2017progressive}, LSGM~\cite{vahdat2021score}, UDM~\cite{kim2022soft}, CIPS~\cite{anokhin2021image}, StyleGAN-v2~\cite{karras2020analyzing}, ImageBART~\cite{esser2021imagebart}, DDPM~\cite{ho2020denoising}, StyleGAN~\cite{karras2019style}, ADM~\cite{dhariwal2021diffusion}, ProjectedGAN~\cite{sauer2021projected}, and LDM~\cite{rombach2022high}. To this end, we train our model on CelebA-HQ, LSUN-bed, and Landscape datasets, and evaluate these models with FID, Precision and Recall metrics to measure both the generated image quality and the coverage of the modeling data manifold. Table~\ref{exp:CelebA-HQ}, \ref{exp:bed}, and \ref{exp:landscape} summary the results on CelebA-HQ, LSUN-beds, and Landscape, respectively. As can be seen in this table, we achieve comparable results in terms of image quality metrics (FID and precision). As for the coverage of the data manifold, we find that our model equips the diffusion model to learn the data manifold with high coverage, resulting in a high Recall rate. We contribute this to the feature pyramid design in the diffusion model, which leverages a  coarse-to-fine learning strategy and allows DM to learn the dataset distribution easier. Moreover, we provide examples generated by \our on these datasets, and the results on CelebA, Landscape, and LSUN-bed are shown in Figure~\ref{demo:celeba},~\ref{demo:landscape}, and~\ref{demo:lsun_bed}, respectively.

\begin{table}[t]
\centering

\resizebox{0.45\textwidth}{!}{
\setlength{\tabcolsep}{1.0mm}{
\begin{tabular}{l|c|c|c}
\multirow{2}{*}{Method} & \multirow{2}{*}{FLOPs} & \multirow{2}{*}{Params} & \multirow{2}{*}{\begin{tabular}[c]{@{}c@{}}Inference time\\ (sec/img)\end{tabular}} \\ 
 &  &  &  \\ \hline
LDM-8 &  37.1 G & 589.8 M & 0.82547 \\ \hline
Baseline &  37.3 G & 1.179 B & 1.02865 \\
+ PyU-Net &  31.7 G & 589.8 M & 0.75052 \\
+ Coarse-to-fine modulation  & 39.7 G & 697.8 M & 0.91782
\end{tabular}}}
\caption{Model ablation of T2I on COCO. The inference time is estimated on single V100 with 32 batch-size. Also, the total inference timesteps are 200s for both LDM and \our.} 
\label{exp:model_cost}
\end{table}

\subsection{Model Analysis}

\subsubsection{MS-VQGAN Performance}
Vector quantization is the first stage of latent diffusion. Its performance may set the upper bound for the final generation quality. The purpose of the quantization model is to encode an image into low-dimensional and discrete space while being able to reconstruct the image. Therefore, we evaluate the reconstructed images of our MS-VQGAN in terms of reconstruction FID (rFID)\footnote{The FID score between the original and reconstructed images.}, IS, PSNR, and SSIM to test the quantization performance.
Also, we compare the results with the state-of-the-art quantization models, VQGAN and VQGAN-kl, and the results are shown in Table~\ref{exp:vqgan}.
For a fair comparison, the codebook sizes (vocabulary size $\times$ embedding dimension) for each setting is set equal.
In Table~\ref{exp:vqgan}, our MS-VQGAN outperforms previous methods on PSNR and SSIM, showing that our quantization model can better preserve visual details.
Meanwhile, our method achieves comparable results on rFID and IS, demonstrating the image fidelity.
Another observation is that keep increasing number of feature scales may not improve the performance, which will be further studied next.

\begin{table}[t]
\center
\resizebox{0.48\textwidth}{!}{
\setlength{\tabcolsep}{1.0mm}{
\begin{tabular}{l|l|cc|cc}
Name & Scales & rFID & IS & PSNR & SSIM \\
\hline
VQGAN$^\dagger$ & f8 & 5.51 & 28.98 & 43.56 & 0.9554 \\
VQGAN-kl$^\dagger$ & f8 & 5.02 & \textbf{31.12} & 42.94 & 0.9548 \\
MS-VQGAN & f16f8 & \textbf{4.67} & \ul{29.31} & \textbf{44.77} & \textbf{0.9664} \\
MS-VQGAN & f32f16f8 & 6.68 & 27.67 & 44.40 & 0.9631 \\

\end{tabular}}}
\vspace{-2mm}
\caption{Performance comparison of quantization models. These models are trained on OpenImage 256*256 images, and evaluated on COCO-stuff valid set. $^\dagger$: obtained from official model checkpoints.}
\vspace{-4mm}
\label{exp:vqgan}
\end{table}

\subsubsection{Feature Scales of \our}
We also analyze the performance of \our with different feature scales (i.e., $N$ in Section~\ref{method:frido}) by training \our on COCO T2I. The results are reported in Table~\ref{exp:scale_analysis}. For $N=3$, we consider \our-f32f16f8 and \our-f16f8f4, and one can find that although the performance on all metrics is improved by using a higher resolution of features (i.e., \our-f16f8f4), the parameter and inference cost also increase accordingly.
This trade-off can also be seen in the $N=2$ settings, where \our-f8f4 boosts the FID from 40.14 to 38.68 while increasing the inference time by $\approx 4$ times (since the resolution of f8f4 is also $\approx 4$ times larger than f16f8).
Moreover, by comparing \our-f16f8 and \our-f32f16f8, we notice that the performance decreases after adding a higher-level feature (i.e., f32).
We hypothesize this is due to the weaker reconstruction ability of MS-VQGAN at f32f16f8 shown in Table~\ref{exp:vqgan}.
The increased FID indicates that adding the f32 layer does not help representing the original image in the quantized code, where the high-resolution codebook's capacity is reduced by 1/3.
Moreover, \our denoising starts from the f32 map, and the low quality of this denoised map results in error accumulation early in the coarse-to-fine generation, hence hurting the denoising at later stages.
With the above observations, we adopt $N=2$ (i.e., \our-f16f8 and \our-f8f4) for all of the tasks.

\begin{table}[t]
\centering
\resizebox{0.47\textwidth}{!}{
\setlength{\tabcolsep}{1.0mm}{
\begin{tabular}{c|ccc|cc}
Name & FID$\downarrow$ & IS$\uparrow$ & CLIP$\uparrow$ & Total Params & \begin{tabular}[c]{@{}c@{}}Inference time\\ (1*V100) 32 BS\end{tabular} \\ \hline
Frido-f32f16f8 &  48.04 & 12.45 & 0.567 & 788.9 M & 1.227 \\
Frido-f16f8f4  & 44.35 & 13.40 & 0.594 & 828.1 M & 6.419 \\
\hline
Frido-f16f8 & 40.14 & 14.25 & \textbf{0.610} & 697.8 M & 0.918 \\
Frido-f8f4 & \textbf{38.68} & \textbf{14.38} & 0.603 & 760.9 M & 3.857
\end{tabular}}}
\vspace{-2mm}
\caption{Analysis of different feature scales on COCO T2I. The inference time is computed on a single V100 with 32 batch-size and 200 inference steps. Note that we compute the total parameters by summing parameters from conditioning transformer, and \our.} 
\vspace{-4mm}
\label{exp:scale_analysis}
\end{table}

\begin{figure}[t]
  \centering
  \includegraphics[page=9
  ,trim={120 92 130 95}, clip, width=0.45\textwidth]{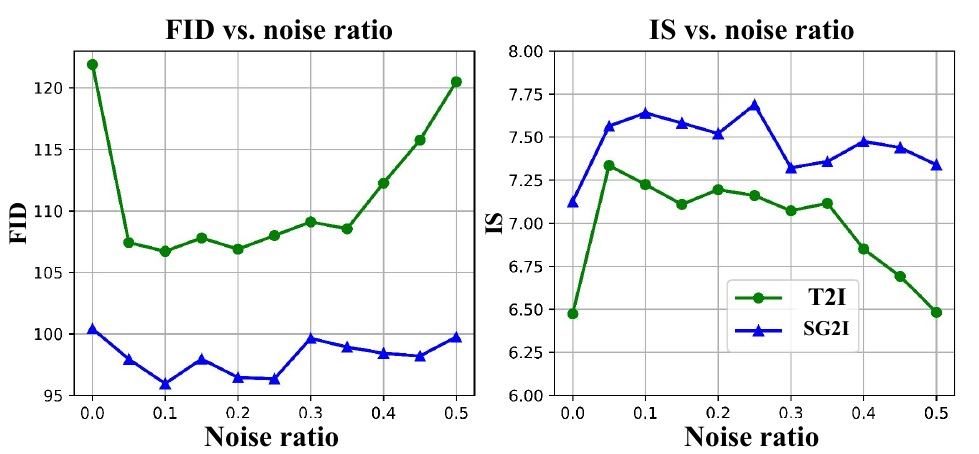}
  \vspace{-2mm}
  \caption{Analysis of noise augmentation.}
  \vspace{-4mm}
  \label{exp:ablation_noise}
\end{figure}

\subsubsection{Noise Augmentation} 
As mentioned in the CFM subsection in Methodology, we use ground truth higher-level feature maps in the training phase similar to teacher forcing in auto-regressive modelling.
Without noise augmentation, the denoising process is prone to overfitting on such ground truth feature.
We suspect this issue is caused by the train-test mismatch, where at inference, the denoising of f8 takes a \emph{model-generated} f16 feature map as the input to the modulation module.
Therefore, the affected f8 feature output results in the lowered quality of visual details in the generated images.
Here we study the effectiveness of the noise augmentation by analyzing different noise ratios $\alpha$ from 0.0 to 0.5. The results are shown in Figure~\ref{exp:ablation_noise}.
We report the FID and IS scores of \our on COCO T2I and VG SG2I. For simplicity, each setting is trained for 220k iterations with 32 batch sizes and evaluated on COCO 2017 validation set and VG validation set for T2I and SG2I, respectively. As shown in this figure, the performance of models without noise augmentation ($\alpha = 0$) increases from 106 to 122 on the FID score of the T2I task.
On the other hand, the performance also drops when the noise ratio is set too high (e.g., $\alpha=0.5$), because the higher-level feature may be too noisy to provide useful information for learning.
By analyzing the trade-off, we choose $\alpha = 0.1$ as a default for \our. 
As can be seen in Figure~\ref{exp:ablation_noise}, this value is quite robust.

\subsubsection{Computation Cost}
Continued from Section~\textcolor{blue}{4.3}, we report the computation cost for each ablated version of \our and compare it with the previous latent diffusion model, LDM~\cite{rombach2022high}, in terms of FLOPs, parameter counts, and inference time. As shown in Table~\ref{exp:model_cost} and Figure~\textcolor{black}{6} in the main paper, we demonstrate that \our is able to boost the performance for all metrics with only a minor increase in parameter count (total of 697M). As for the inference time, we set the total timesteps to 200s for both LDM and \our and report the inference time (sec) per image. In this table, we can find that under the same inference timestep setting, \our achieve comparable inference cost per image compared with LDM, while as shown in Figure~\textcolor{black}{7}, \our achieves decent performance improvement on both tasks. This again verifies the efficiency of \our in high-quality conditional image generation.

\section{Additional Discussion}

\subsection{Limitations and Future Directions}

We observed that the distribution of the encoded features (by MS-VQGAN) at each scale is not regularized, resulting in large variations in the mean and standard deviation of features at different scales. This could further bring negative effects on the learning of the diffusion model. Specifically, in the diffusion process, the noisy data are created by interpolating input features with standard normal variations, requiring the distribution of input features to better be with mean$\approx$0 and std$\approx$1. Therefore, the non-regularized input features may not be corrupted into noise in the diffusion process as expected, damaging the learning of the denoising process. In this paper, we mitigate this problem by scaling each scale of feature independently with the reciprocal of the standard deviation of the corresponding feature. However, we believe that a regularization objective is needed to constrain the distribution of the encoded latent in such a quantization process for the diffusion model.

Another potential future direction is to explore the question of ``what kind of high or low knowledge encoded in the features would benefit the coarse-to-fine learning of the diffusion model?". In \our, we design multi-scale VQGAN with a feature pyramid fusion module, enabling us to encode an input image into features with multiple scales and extract high and low-level information implicitly. To explore the aforementioned question, we may enhance MS-VQGAN by adding an objective function to guide the information extracted on high or low-level features. Specifically, a reconstruction loss between the low resolution of the input image and the reconstructed image decoded by only high-level features may be imposed to MS-VQGAN to guide the multi-scale quantization explicitly. We will leave these directions for future research.

\subsection{Ethics Concerns}

Similar to other generative works, \our would be also vulnerable to \textit{malicious} use because of its powerful ability to generate real-world images based on complex cross-modal conditions, ranging from the label, natural language, bounding box layout, and scene graph. Moreover, the \textit{fairness} for generating particular categories of objects (e.g., human) would be a potential problem for generative models used in a real-world scenario. We contribute such social bias to the potential data collecting bias in the training datasets. Such potentially social issues had been identified and discussed in the recent works~\cite{tan2020improving}, and some possible solutions to mitigate these problems are considered in a survey~\cite{brundage2018malicious}.

\begin{figure*}[!th]
  \centering
  \includegraphics[page=3,trim={450 320 450 320}, clip, width=\textwidth]{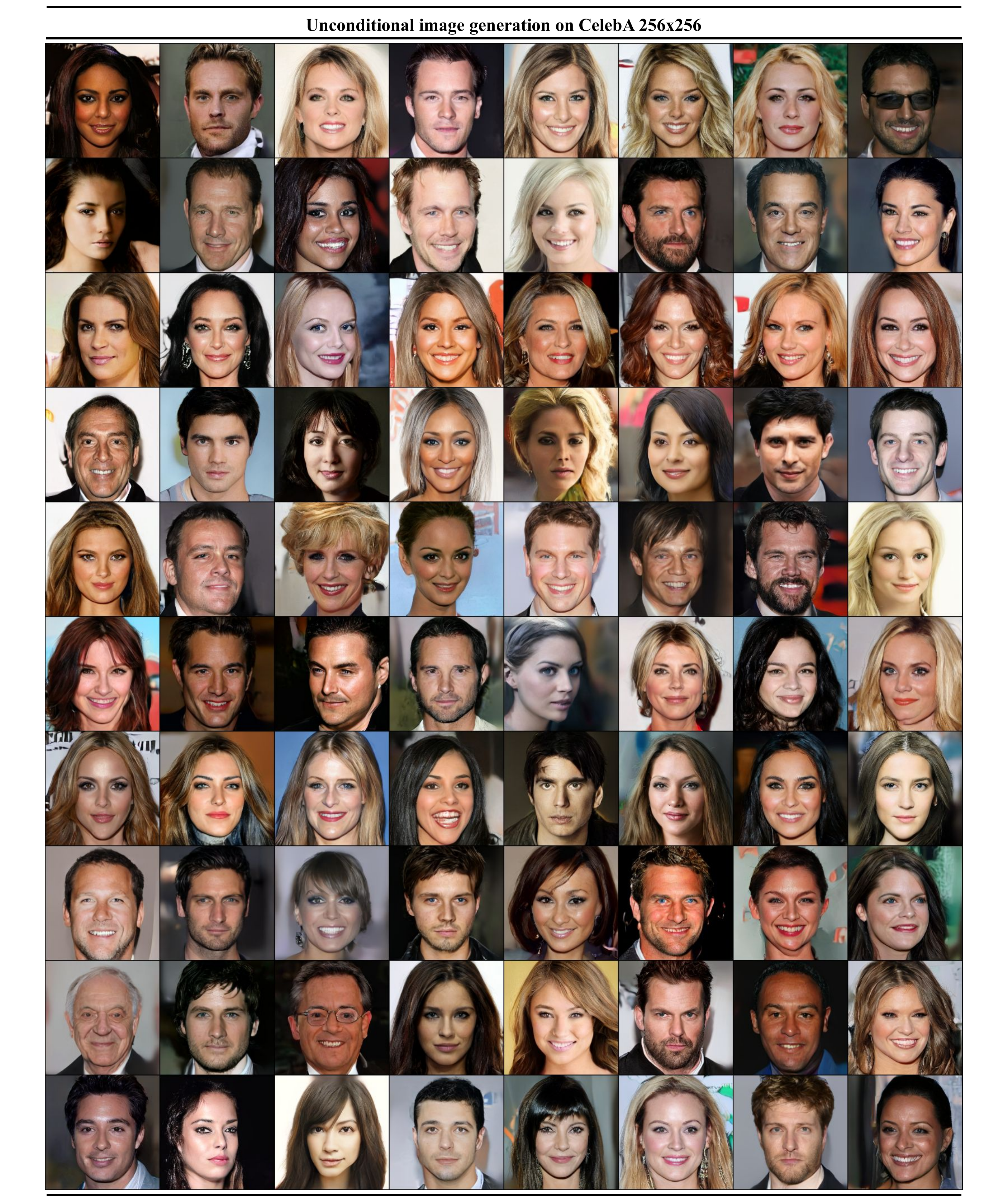}
  \vspace{-4mm}
  \caption{Generated examples of unconditional image generation on CelebA 256x256.}
  \vspace{-2mm}
  \label{demo:celeba}
\end{figure*}

\begin{figure*}[!th]
  \centering
  \includegraphics[page=1,trim={450 320 450 320}, clip, width=\textwidth]{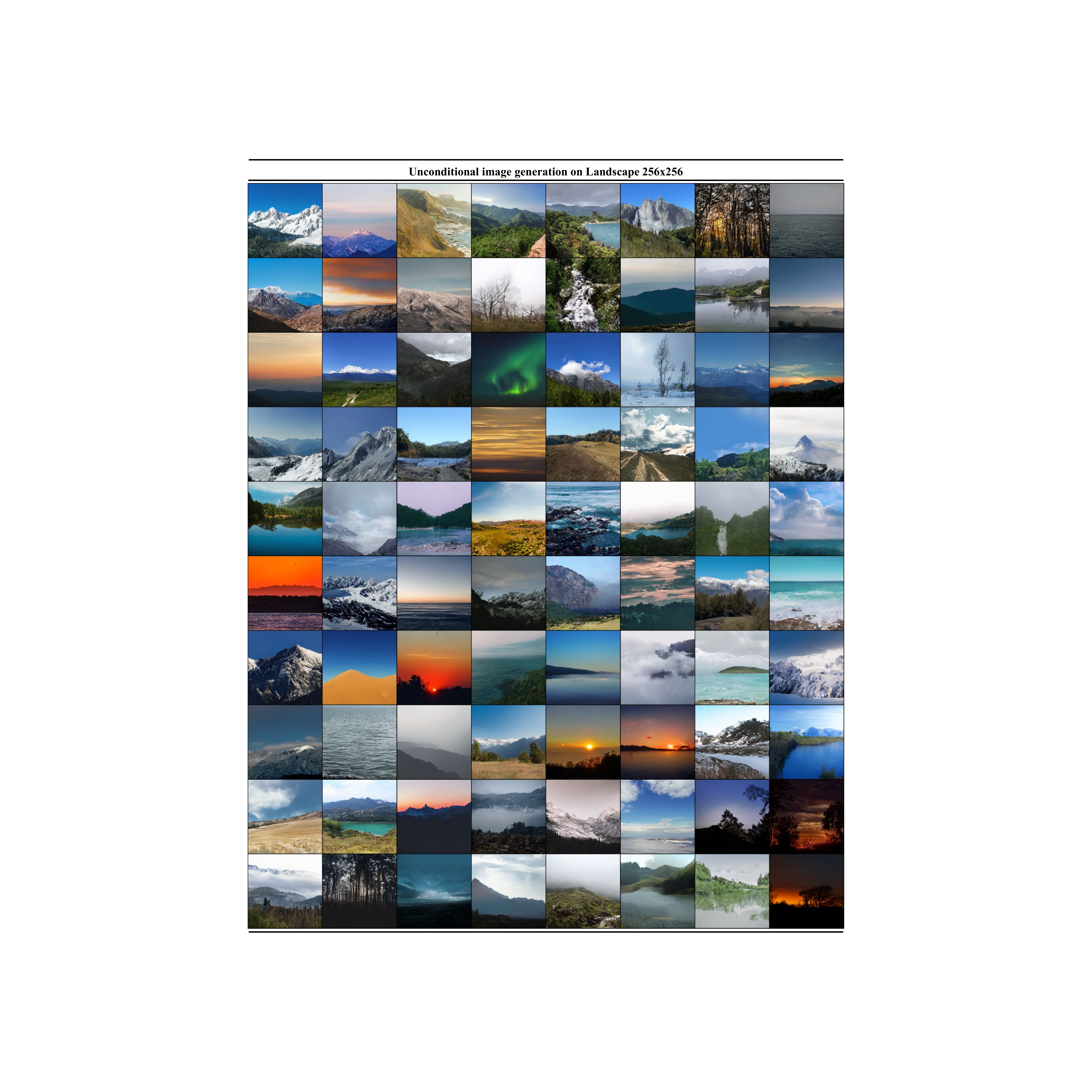}
  \vspace{-4mm}
  \caption{Generated examples of unconditional image generation on Landscape 256x256.}
  \vspace{-2mm}
  \label{demo:landscape}
\end{figure*}

\begin{figure*}[!th]
  \centering
  \includegraphics[page=2,trim={450 320 450 320}, clip, width=\textwidth]{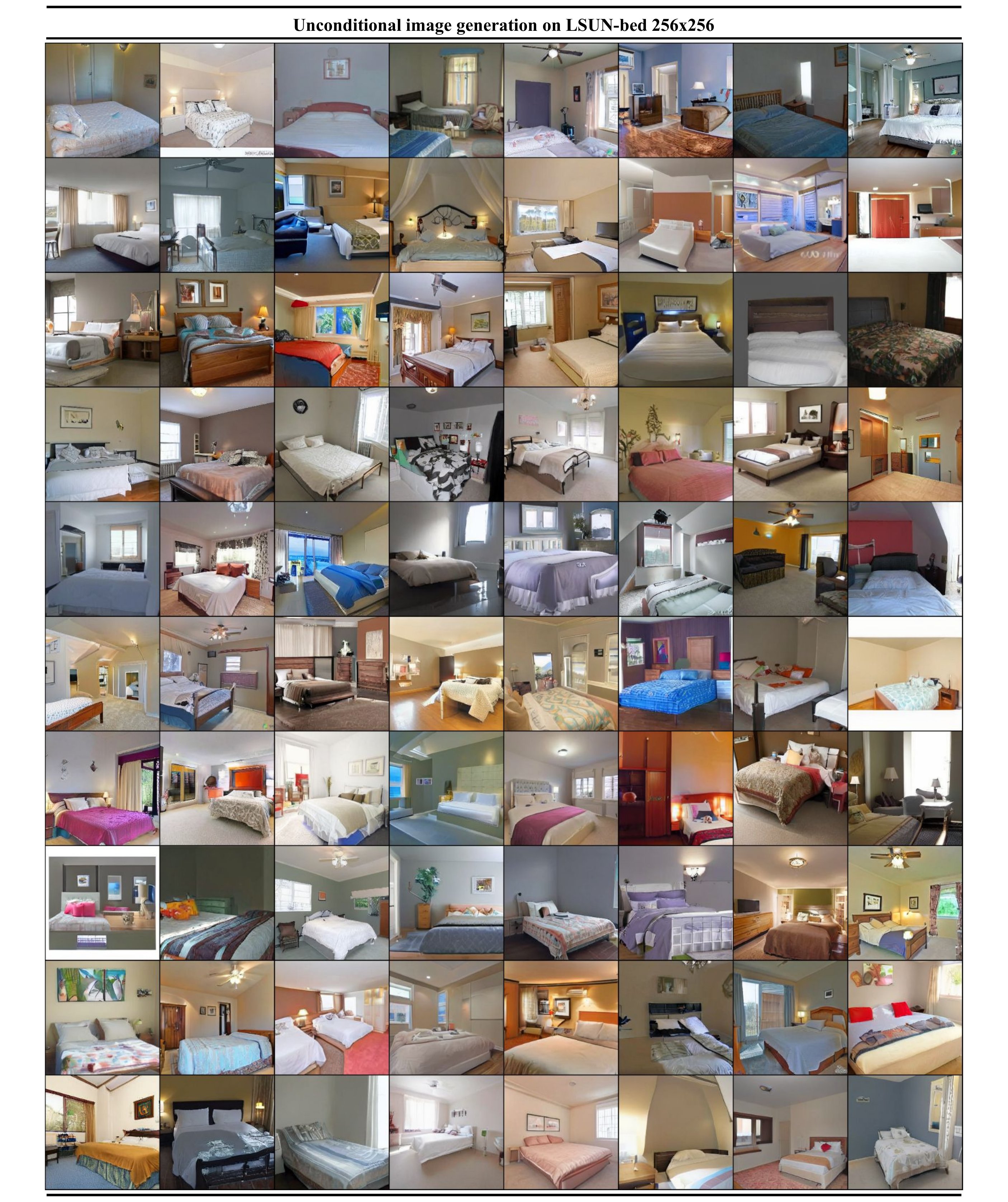}
  \vspace{-4mm}
  \caption{Generated examples of unconditional image generation on LSUN-bed 256x256.}
  \vspace{-2mm}
  \label{demo:lsun_bed}
\end{figure*}

\end{document}